\documentclass[sigconf]{acmart}

\usepackage{graphicx}
\usepackage{amsmath}

\usepackage{amssymb}
\usepackage{booktabs}
\usepackage{multirow}
\usepackage{algorithm}
\usepackage{algorithmic}
\usepackage{caption}
\usepackage{subcaption}
\usepackage{pifont}
\usepackage{balance}

\definecolor{lightblue}{RGB}{143,176,220}
\definecolor{lightgold}{RGB}{255,210,70}

\AtBeginDocument{%
  \providecommand\BibTeX{{%
    \normalfont B\kern-0.5em{\scshape i\kern-0.25em b}\kern-0.8em\TeX}}}

\copyrightyear{2022}
\acmYear{2022}
\setcopyright{acmlicensed}
\acmConference[MM '22]{Proceedings of the 30th ACM International Conference on Multimedia}{October 10--14, 2022}{Lisboa, Portugal}
\acmBooktitle{Proceedings of the 30th ACM International Conference on Multimedia (MM '22), October 10--14, 2022, Lisboa, Portugal}
\acmPrice{15.00}
\acmDOI{10.1145/3503161.3548163}
\acmISBN{978-1-4503-9203-7/22/10}

\acmSubmissionID{1700}

\begin{document}

\title{Neighbor Correspondence Matching for Flow-based Video Frame Synthesis}


\author{Zhaoyang Jia}
\authornote{This work was done when Zhaoyang Jia was an intern at Microsoft Research Asia.}
\email{jzy_ustc@mail.ustc.edu.cn}
\affiliation{%
  \institution{University of Science and Technology of China}
  \streetaddress{No. 443, Huangshan Road}
  \city{Hefei}
  \state{Anhui}
  \country{China}
  \postcode{230027}
}

\author{Yan Lu}
\email{yanlu@microsoft.com}
\affiliation{%
  \institution{Microsoft Research China}
  \streetaddress{No. 5 Dan Ling Street}
  \city{Beijing}
  \country{China}
  \postcode{100080}
}

\author{Houqiang Li}
\email{lihq@ustc.edu.cn}
\affiliation{%
  \institution{University of Science and Technology of China}
  \streetaddress{No. 443, Huangshan Road}
  \city{Hefei}
  \state{Anhui}
  \country{China}
  \postcode{230027}
}


\renewcommand{\shortauthors}{Zhaoyang Jia et al.}

\begin{abstract}
Video frame synthesis, which consists of \textit{interpolation} and \textit{extrapolation}, is an essential video processing technique that can be applied to various scenarios. However, most existing methods cannot handle small objects or large motion well, especially in high-resolution videos such as 4K videos. To eliminate such limitations, we introduce a \textbf{n}eighbor \textbf{c}orrespondence \textbf{m}atching (\textbf{NCM}) algorithm for flow-based frame synthesis. Since the current frame is not available in video frame synthesis, NCM is performed in a current-frame-agnostic fashion to establish multi-scale correspondences in the spatial-temporal neighborhoods of each pixel. Based on the powerful motion representation capability of NCM, we further propose to estimate intermediate flows for frame synthesis in a heterogeneous coarse-to-fine scheme. Specifically, the coarse-scale module is designed to leverage neighbor correspondences to capture large motion, while the fine-scale module is more computationally efficient to speed up the estimation process. Both modules are trained progressively to eliminate the resolution gap between training dataset and real-world videos. Experimental results show that NCM achieves state-of-the-art performance on several benchmarks. In addition, NCM can be applied to various practical scenarios such as video compression to achieve better performance.
\end{abstract}



\begin{teaserfigure}
    \includegraphics[width=\linewidth]{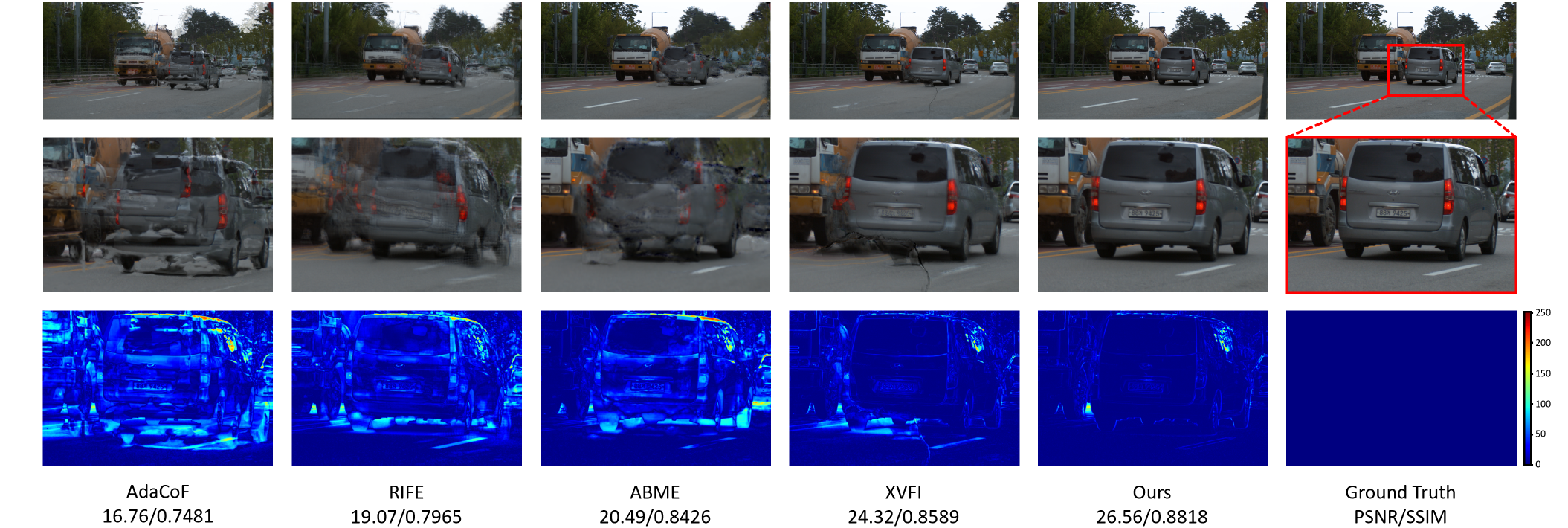}
    \caption{A difficult example of video frame interpolation in X4K1000FPS\cite{sim2021xvfi}. From top to bottom are the synthesised results, the zoomed details, and the residuals between synthesised images and ground truth. Previous methods fail on the car with large motion and produce blurs or flickers, while our method can generate high-quality results.}
    \label{fig1}
\end{teaserfigure}


\settopmatter{printacmref=false} 
\renewcommand\footnotetextcopyrightpermission[1]{} 
\pagestyle{plain} 

\maketitle

\begin{figure*}
    \includegraphics[width=0.75\linewidth]{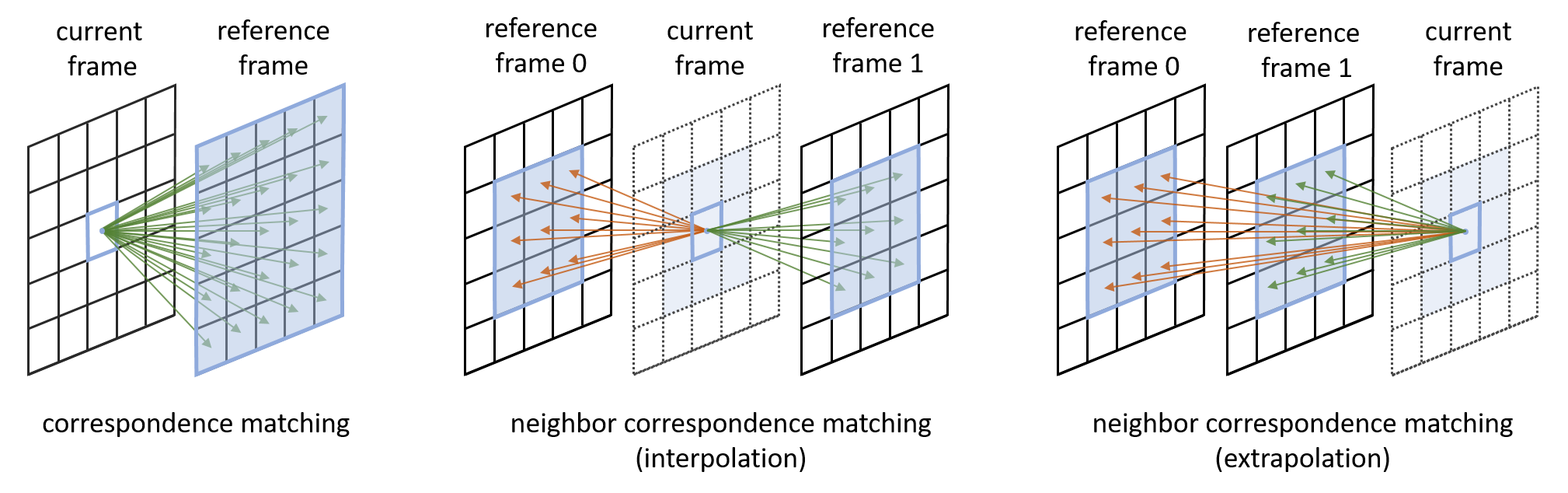}
    \caption{Illustration of neighbor correspondence matching. \textcolor{lightblue}{Blue} regions in the reference frames denote the matching regions. The correspondence matching \cite{cheng2021rethinking,teed2020raft} is performed between the current frame and the reference frame, while our neighbor correspondence matching is performed in a current-frame-agnostic fashion to match pixels in the spatial-temporal neighborhoods of the current frame.}
    \label{fig2}
\end{figure*}

\section{Introduction}
\label{sec:intro}

Video frame synthesis is a classic video processing task to generate frame in-between (interpolation) or subsequent to (extrapolation) reference frames. It can be applied to many practice applications, such as video compression \cite{wu2018video, pourreza2021extending}, video view synthesis \cite{kalantari2016learning, flynn2016deepstereo}, slow-motion generation \cite{jiang2018super} and motion blur synthesis \cite{brooks2019learning}. Recently, various deep-learning-based algorithms have been proposed to handle video frame synthesis problem. Most of them focus on interpolation \cite{jiang2018super, bao2019memc,park2020bmbc, huang2020rife,park2021asymmetric, lee2020adacof, niklaus2017video, sim2021xvfi}, while some others \cite{liu2017video} can deal with both interpolation and extrapolation in a unified framework.

Among existing video frame synthesis algorithms, flow-based schemes predict the current frame by warping reference frames with estimated optical flows. Many flow-based interpolation schemes \cite{jiang2018super, bao2019memc, bao2019depth} first compute bi-directional optical flows, and then approximate intermediate flows by flow reversal \cite{jiang2018super, liu2020enhanced}. On the contrary, some recent schemes \cite{park2020bmbc,park2021asymmetric,huang2020rife} directly estimate intermediate flows and achieve superior performance. However, most of existing methods fail in estimating large motion or motion of small objects, as shown in Fig.\ref{fig1}. It is mainly caused by the limited receptive field and motion capture capability of CNNs.

Correspondence matching is proven to be effective in capturing long-term correlations in multimedia tasks like video object segmentation \cite{yang2021collaborative,cheng2021rethinking} and optical flow estimation \cite{teed2020raft,jiang2021learning}. In these scenarios, by matching pixels of the current frame in the reference frame, a correspondence matrix can be established to guide the generation of mask or flow (Fig.\ref{fig2}, left). However, in video frame synthesis, we only have two inference frames and the current frame is not available. As a result, the correspondence matching cannot be performed directly. So \textit{how to perform correspondence matching in video frame synthesis} is still an unanswered question.

In this paper, we introduce a \textbf{n}eighbor \textbf{c}orrespondence \textbf{m}atching (\textbf{NCM}) algorithm to enhance flow estimation in video frame synthesis, which can establish correspondences in a current-frame-agnostic fashion. Observing that objects usually move continuously and locally within a small region in natural videos, we propose to perform correspondence matching between the spatial-temporal neighbors of each pixel. Specifically, for each pixel in the current frame, we only use pixels in the local windows of adjacent reference frames to calculate the correspondence matrix  (Fig.\ref{fig2}, middle and right), so the current frame is not required in this process.
The matched neighbor correspondence matrix can effectively model the object correlations, from which we can infer sufficient motion cues to guide the generation of flows. In addition, multi-scale neighbor correspondence matching is further preformed to extend the receptive field and capture large motion. 

Compared with previous cost-volume-based schemes \cite{sun2018pwc,teed2020raft, park2020bmbc} that guide the matching region by estimated flow, NCM works better on video frame synthesis. Due to the complexity of frame synthesis, the estimated flows are usually inaccurate at the beginning of estimation. In this case, if the matching region is guided by the inaccurate flow, the matching region may move away from where the current pixel is, leading to ineffective matching. On the contrary, NCM directly matches correspondences in fixed neighbor regions to avoid misleading of inaccurate flows, which is more stable and more efficient since the correspondences only need to be computed once to be applied in different stages of estimation.


Based on NCM, we further propose a unified video frame synthesis network for both interpolation and extrapolation. The proposed model can accurately estimate intermediate flows in a heterogeneous coarse-to-fine scheme. Specifically, the coarse-scale module is designed to utilize multi-scale neighbor correspondence matrix to capture accurate motion, while the fine-scale module refines the coarse flows in a computationally efficient fashion. With the proposed heterogeneous coarse-to-fine structure, our model is not only effective but also efficient, especially for high-resolution videos.

For flow-based video frame synthesis schemes, another existing problem is the resolution gap between training dataset and real-world high-resolution videos. To eliminate such gap, we propose to train coarse and fine-scale modules using a progressive training strategy. Combining all above designs, we can augment RIFE \cite{jiang2021learning} framework to a novel NCM-based network, which demonstrates new state-of-the-art results in several video frame synthesis benchmarks. Specifically, In challenging X4F1000FPS benchmark, our model improves PSNR by 1.47dB (from 30.16dB of ABME  \cite{park2021asymmetric} to 31.63dB), which shows its capability in capturing large motion and handling real-scenario videos.

NCM can be extended to many practical applications such as video compression, motion effects generation, and jitters removal in real-time communication. We build a video compression algorithm based on NCM, and results show out model can save 10\% bits in HEVC dataset compared with H265-HM \cite{HM}.

In summary, the main contributions in this paper are : 
\begin{enumerate}
\item We introduce a neighbor correspondence matching algorithm for video frame synthesis, which is simple yet effective in capturing large motion or small objects.
\item We propose a heterogeneous coarse-to-fine structure, which can generate intermediate flows both accurately and efficiently. We further train them in a progressive fashion to eliminate the resolution gap between training and inference.
\item Combining all designs above, we propose a unified framework for video frame synthesis. It achieves the new state-of-the-art in both interpolation and extrapolation, which improves PSNR by 1.47dB (30.16dB$\to$31.63dB) compared with previous SOTA ABME
\cite{park2021asymmetric} on X4K1000FPS.
\end{enumerate}

\section{Related Works}
\label{sec:rela}

\begin{figure*}
    \includegraphics[width=\linewidth]{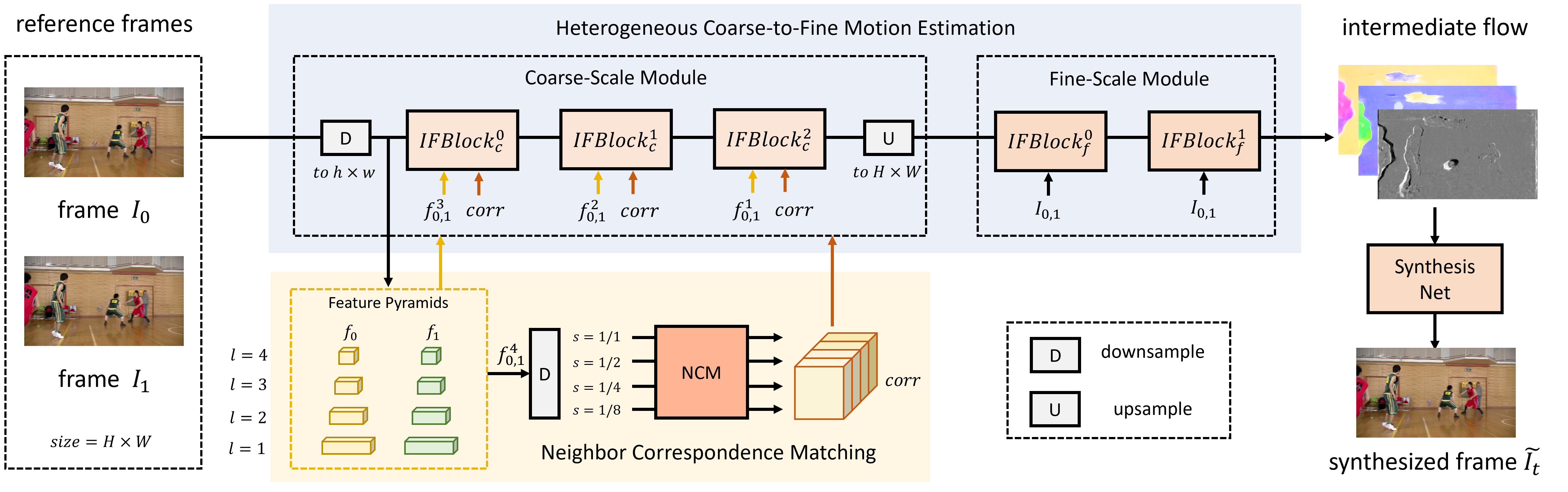}
    \caption{Overview of the proposed network. Our network is based on RIFE \cite{huang2020rife}. We augment RIFE with : 1) the neighbor correspondence matching (NCM, \textcolor{lightgold}{yellow} color) with a feature pyramid, 2) the heterogeneous coarse-to-fine modules (\textcolor{lightblue}{blue} color). IFBlocks estimate flows with features $f_{0,1}^l$, correspondences $corr$ or frame $I_{0,1}$ as inputs, which will be illustrated in Fig.\ref{fig5}. }
    \label{fig3}
\end{figure*}

\subsection{Video Frame Interpolation}
Video frame interpolation (VFI) is a sub-task of video frame synthesis, which aims to predict the intermediate frame between input frames. Learning-based VFI methods can be categorised as kernel-based methods \cite{niklaus2017video, lee2020adacof} and flow-based methods \cite{liu2017video, jiang2018super, bao2019memc, park2020bmbc, park2021asymmetric, huang2020rife}. Kernel-based VFI learns motion implicitly using dynamic kernels \cite{niklaus2017video} and deformable kernels \cite{lee2020adacof}, which can preserve structural stability but might generate blurry frames because of the lack of explicit motion guidance. On the contrary, flow-based VFI explicitly model the motion with dense pixel-wise flows and perform forward-warp \cite{niklaus2018context,niklaus2020softmax} or backward-warp \cite{jiang2018super,park2020bmbc, park2021asymmetric, huang2020rife} to predict the frame, which can achieve superior performance. Since forward-warping can cause holes and overlaps in the warped image, backward-warping is more widely exploited and applied in flow-based VFI.

For flow-based VFI that perform backward-warping, the key is \textit{how to estimate the intermediate flows}. The intermediate flows should be spatially aligned with the current synthesised frame, but such spatial information is agnostic during inference. That makes it difficult to estimate accurate intermediate flows. Early flow-based VFI leverage advanced optical flow methods \cite{ranjan2017optical, sun2018pwc, teed2020raft, jiang2021learning} to estimate bi-directional flows, and perform flow reversal to generate intermediate flows. Later, Park et al. \cite{park2020bmbc} estimates symmetric bilateral motion with a bilateral cost
volume, which is further improved by Park et al. \cite{park2021asymmetric} through introducing asymmetric motion to achieve superior performance. Recently, Huang et al. \cite{huang2020rife} proposed to estimate intermediate flows directly with a privileged distillation supervision, which shows a new paradigm for intermediate flow estimation. However, these schemes cannot handle large motion of small objects well, and are limited by the solution gap between training and inference. It inspires us to explore more effective motion representations for intermediate flow estimation.

\subsection{Video Frame Extrapolation}
Video frame extrapolation aims to predict the frame subsequent to input frames. It is much more challenging than interpolation because unseen objects may exist in the current frame. Liu et al. \cite{liu2017video} proposed a unified framework for both interpolation and extrapolation, which models intermediate flow as a 3D voxel flow and synthesises current frame by trilinear sampling.
However, due to the difficulty of synthesis frame only by two previous frames, many following works focus more on multi-frame extrapolation or video prediction \cite{lee2021video,le2021ccvs,wang2021predrnn}. These works can generate more accurate results, but they usually need a sequence of frames to warm up, which are computationally expensive. In this paper, our synthesis algorithm is more like Liu et al. \cite{liu2017video}, that only needs two frames as input and can adapt to both interpolation and extrapolation.

\subsection{Correspondence Matching}

Correspondence matching is a technique to establish correspondences between images, which has been widely used in many computer vision and graphics tasks. In many 3D vision tasks \cite{hartley2003multiple,agarwal2011building,schonberger2016structure}, correspondences are computed between different views to explore the 3D structure. In video object segmentation  \cite{yang2021collaborative,cheng2021rethinking}, correspondence matching is performed to search the similar pixels in the reference frames to propagate the mask. Benefiting from the long-term correlation modeling capability of correspondence matching, these schemes achieve remarkable performance.

Recently, correspondence is leveraged in flow estimation \cite{teed2020raft, park2020bmbc} and achieve superior performance. RAFT \cite{teed2020raft} builds an all-pair correspondence matrix and looks up it to refine estimated optical flow recurrently, but it cannot be effectively applied in video frame synthesis because the current frame is not available to compute correspondences. BMBC \cite{park2020bmbc} establishes a bilateral cost volume in video frame interpolation, but it is limited by the symmetric linear motion assumption. In addition, these schemes introduce correspondence as a means to refine estimated flows, which can be easily misled if inaccurate flows are given. In this paper, we rethink the correspondence matching in flow estimation. Based on the assumption that objects usually move continuously and locally in natural videos, we introduce neighbor correspondence matching as a new manner for motion correlation matching.

\section{Methods}
\label{sec:methods}
The overview of the proposed video frame synthesis network is shown in Fig.\ref{fig3}. The network consists of three parts : 1) neighbor correspondence matching with a feature pyramid (\textcolor{lightgold}{yellow} in Fig.\ref{fig3}), 2) heterogeneous coarse-to-fine motion estimation(\textcolor{lightblue}{blue} in Fig.\ref{fig3}), and 3) frame synthesis. For completeness, we first briefly introduce RIFE \cite{huang2020rife} from which we adopt some block designs, and then demonstrate details of each module in this section.

\subsection{Background}
\label{sec:3-0}
We base our network design on the RIFE \cite{huang2020rife} framework. In RIFE, given a pair of reference frames $I_0, I_1\in\mathbb{R}^{3\times H \times W}$, three \textit{IFBlock}s (Fig.\ref{fig5}, left) are used to estimate intermediate flows from the coarse to fine-scale. The flows $F=(F_{t\to0}, F_{t\to1})$ and fusion map $M$ are refined by residual estimation in each IFBlock, and the current image at time $t$ can be generated by:
\begin{equation}
    \hat{I_t} = M\odot warp(I_0, F_{t\to0}) + (1-M)\odot warp(I_1, F_{t\to1})
\end{equation}
where $\odot$ denotes pixel-wise product and $warp$ means backward warping operation. Then $I_t$, $F$ and $M$ are fed to a U-Net-like refine network (i.e., the synthesis network) to generate the synthesised frame $\tilde{I_t}$. 

RIFE is light weight and real-time, but the synthesis quality is not satisfactory due to the limited receptive field and motion capture capability of the designed fully-convolutional network. In addition, RIFE cannot adapt to high-resolution videos where the motion is even larger. To eliminate these limitations, we propose neighbor correspondence matching and a heterogeneous coarse-to-fine structure for video frame synthesis, which can effectively estimate accurate intermediate flows even on 4K videos.

\subsection{Neighbor Correspondence Matching}
\label{sec:3-1}

\subsubsection{Overview}
Based on the observation that an object usually move continuously and locally within a small region in natural videos, the core idea of NCM is to explore the motion information by establishing spatial-temporal correlation between the neighboring regions. In detail, we compute the correspondences between the local windows of two adjacent reference frames for each pixel, as shown in Fig.\ref{fig2}. It means we need no information about the current frame, so the matching can be performed in a current-frame-agnostic fashion to meet the need of frame synthesis. 

It is worth noting that the position of local windows are determined by the position of pixel, which is different from the cost-volume-based manners \cite{huang2020rife, park2020bmbc} that establish a cost volume around where the estimated flows point. The flow-centric manners have a potential problem that if the estimated flow is inaccurate, the matching may be performed in a wrong region. As a result, the cost volume cannot compute effective correlations to refine the estimated flows. On the contrary, NCM is pixel-centric and will not be misled by the inaccurate flows. Experiments also show that compared with flow-guided matching, NCM can lead to better performance and stability.

\subsubsection{Mathematical Formulation}
Given a pair of reference frames $I_0, I_1\in\mathbb{R}^{3\times H \times W}$, a $n$-layer feature pyramid $f_i^l\in\mathbb{R}^{C_l\times H_l\times W_l}, i\in\{0,1\}$ is first extracted with several residual blocks\cite{he2016deep}, where $l\in\{1, \dots, n\}$ denotes different layers and $C_l, H_l, W_l$ are the channel number, height and width of the feature from the $l$-th layer. The first $n-1$ features only serve as the image features for subsequent modules, while feature from the deepest layer $f^n$ is used for correspondence matching to generate motion feature.

For a pixel at spatial position $(i, j)$, we perform NCM to compute the correspondences in $d\times d$ windows : 
\begin{equation}
    corr^0(i,j) = \{f_0^n(i+\delta_{i_0},j+\delta_{j_0})\cdot f_1^n(i+\delta_{i_1},j+\delta_{j_1})\}_{\delta_{i,j_{0,1}}}
\end{equation}
where $\delta_{i,j_{0,1}}\in\{-d/2, -d/2+1,\dots,d/2\}$ denote different location pairs in the window, and $\cdot$ denotes the channel-wise dot production. The computed correspondence matrix $corr^0\in \mathbb{R}^{d^4\times H_n\times W_n}$ contains correlations of all pairs in the neighborhoods, which can be further leveraged to extract motion information.

\begin{figure}
    \includegraphics[width=\linewidth]{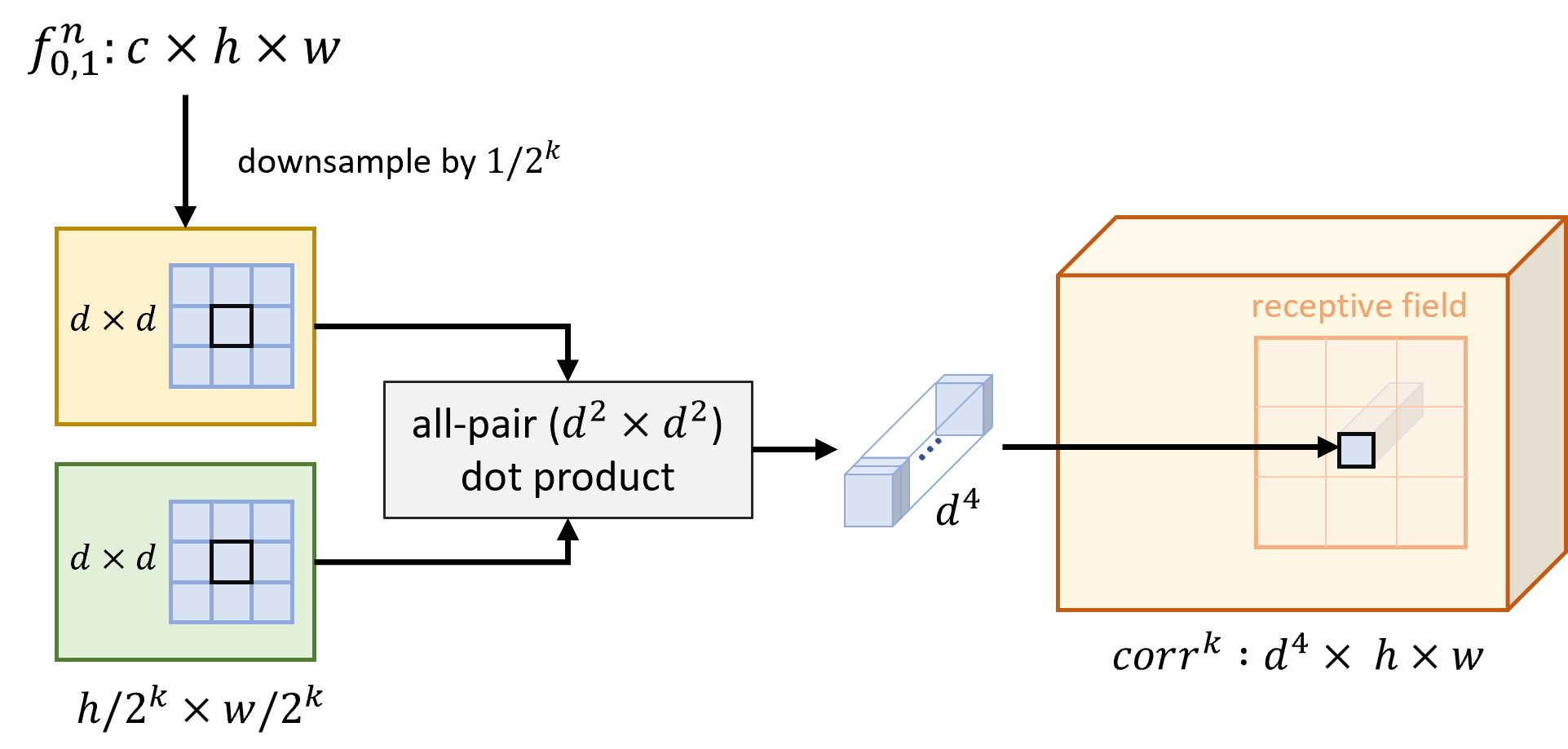}
    \caption{Illustration of neighbor correspondence matching at the $k$-th scale.}
    \label{fig4}
\end{figure}

To enlarge the receptive field and capture large motion, we further perform multi-scale correspondence matching. As shown in Fig.\ref{fig4}, we first downsample $f^n$ to $s=1/2^k$ resolution to generate multi-scale features $f^{n^k}, k\in\{0,1,\dots, K\}$. For each level $k$, the correspondences can be computed by :
\begin{equation}
    corr^k(i,j) = \{f_0^{n^k}(i_k+\delta_{i_0},j_k+\delta_{j_0})\cdot f_1^{n^k}(i_k+\delta_{i_1},j_k+\delta_{j_1})\}_{\delta_{i,j_{0,1}}}
\end{equation}
where $(i_k,j_k) = (i/2^k,j/2^k)$ is the position of pixel in the downsampled feature map. We use bilinear interpolation for non-interger position. And the final multi-scale neighbor correspondences can be generated by simply concatenating correspondences at different levels:
\begin{equation}
    corr = corr^0\ |\ corr^1\ |\ \cdots\ |\ corr^K
\end{equation}
where $|$ denotes channel concatenation. In this paper, we extract $n=4$ layers feature pyramid with $1, 1/2, 1/4, 1/8$ resolutions of input frames, and perform NCM in 4 scales ($K=3$) with window size $d=3$.

\begin{figure}
    \includegraphics[width=\linewidth]{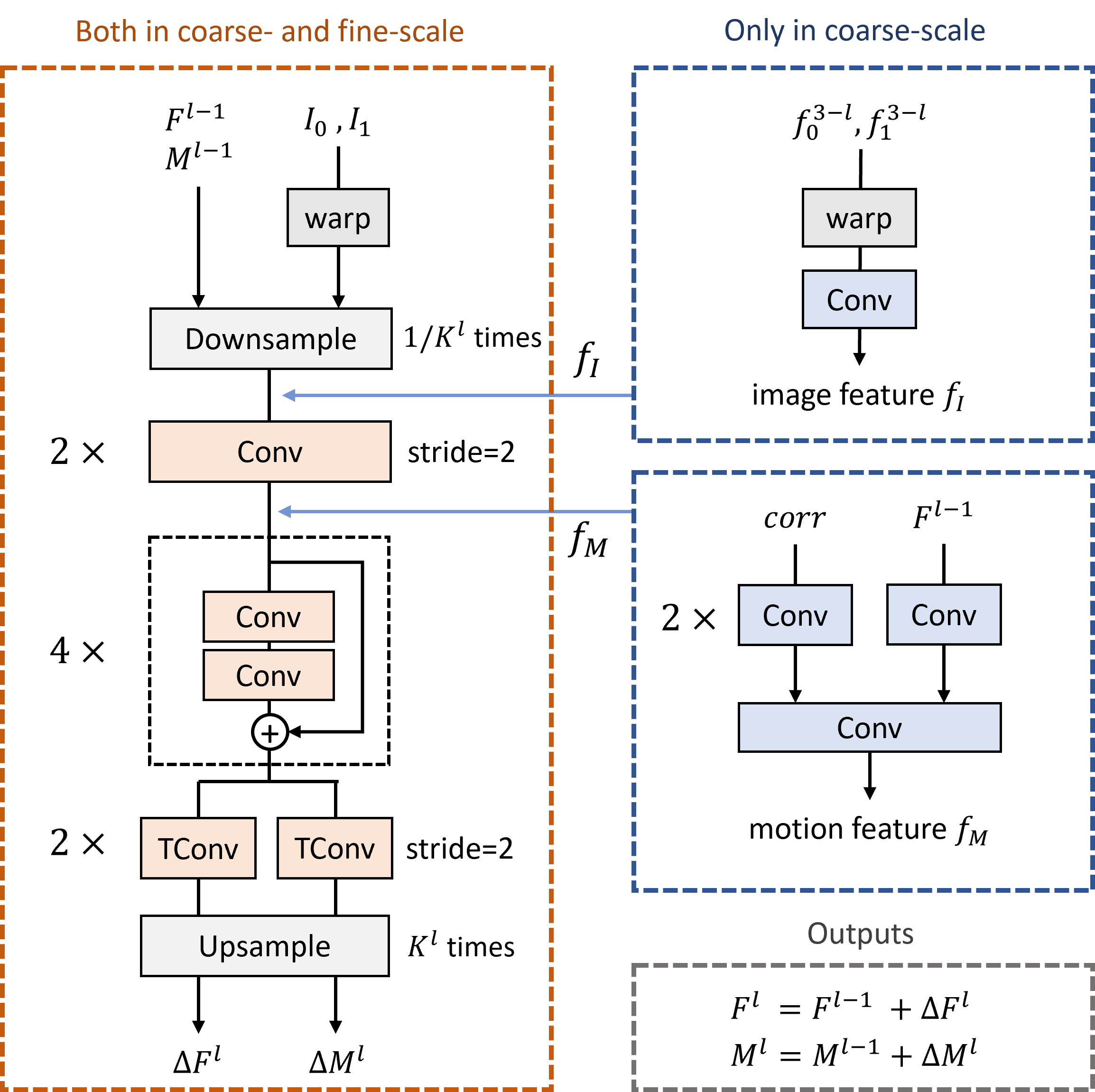}
    \caption{Structure of \textit{IFBlock}. In the left we show the original IFBlock in IFNet-HD \cite{huang2020rife}, which is used as fine-scale \textit{IFBlock}$_f$ in the proposed network. In the right we show our modification to add image features and motion features into IFBlock, which is used only in coarse-scale \textit{IFBlock}$_c$.}
    \label{fig5}
\end{figure}

\subsection{Heterogeneous Coarse-to-Fine flow estimation}
\label{sec:3-2}

Existing coarse-to-fine flow estimation manners \cite{sun2018pwc, park2020bmbc, huang2020rife} usually adopt the same upsampling factor and the same model structure from coarse to fine-scale. However, it may not be the best solution for coarse-to-fine scheme, because the coarse-scale and fine-scale have different focus on motion estimation. In coarse-scale, the flows need to be estimated from scratch, so strong motion capture capability is preferred. In fine-scale, we only need to refine the coarse-scale-flows, which can be done with fewer cost. Based on such idea, we propose a heterogeneous coarse-to-fine structure to adopt different module designs for coarse and fine-scale.

Our heterogeneous coarse-to-fine structure comprises of a coarse-scale module and a fine-scale module. To adapt to different input resolutions, we downsample $I_0, I_1$ to $(h,w)$ to feed into the coarse-scale module, and the value of $(h,w)$ can be decided by the resolution of the input video. The estimated coarse flow is upsampled back to original resolution and fed to the subsequent fine-scale module.

The coarse-scale module is designed to leverage the neighbor correspondences for more accurate flows. In detail, we perform NCM to obtain feature pyramid $f^l$ and neighbor correspondences $corr$, which are fed into three augmented IFBlocks to estimate the coarse-scale flows. As shown in Fig.\ref{fig5} right, in each IFBlock, we warp $f^l$ to generate an image feature $f_I$, and fuse $corr$ and flows for a motion feature $f_M$. The residual flows and mask $\Delta F^l, \Delta M^l$ are estimated to refine that of the previous block : 
\begin{align}
    &F^l=F^{l-1}+\Delta F^l\\
    &M^l=M^{l-1}+\Delta M^l
\end{align}

For the fine-scale module, we directly adopt original IFBlocks to be more computationally efficient. Two IFBlocks receive the fine-scale flows as input, and refine it only using the high-resolution frames. Finally, the estimated intermediate flows are fed into the synthesis network to generate the synthesized frame $\tilde{I_t}$ as output.

The estimation resolution in each IFBlock can be controlled flexibly by the size $(h,w)$ and the downsample factor $K_{c,f}$ to adapt to the resolution of the input video. Assume $H<W$, we use parameter $a$ to control the the size by $(h,w)=(a,W/H\times a)$. In the fine-scale module, we set $K_f=(2,1)$ if $a/H<1/2$ otherwise set $K_f=(1,1)$. We set $K_c = (4,2,1)$ in the coarse-scale module.
 
\subsection{Progressive Learning}
\label{sec:3-3}

Many existing frame synthesis schemes cannot be well extended to applications due to the resolution gap between training and inference. That is, the training data is low-resolution (e.g., $256\times256$) but the resolution of real-world data may be much higher (e.g., 1080p or 4K). To address this problem, we design a progressive learning scheme for the proposed network. The basic idea is to separate the end-to-end training into two stages to simulate the inference on high-resolution videos: 
\begin{itemize}
\item In stage \uppercase\expandafter{\romannumeral1}, only the coarse-scale module is trained on low-resolution $256\times256$ frames. It can be regarded as training on the low-resolution version of real-world high-resolution images.
\item In stage \uppercase\expandafter{\romannumeral2}, the coarse-scale module is fixed, and the fine-scale module is trained to refine the coarse-scale flows to high-resolution. We randomly downsample images to $64\times64$, $128\times128$ or keep $256\times256$ in each mini-batch to generate pseudo low-high resolution data pairs for training.
\end{itemize}
In inference, the coarse-module can estimate accurate low-resolution flows with stage \uppercase\expandafter{\romannumeral1}, and the fine-scale module can refine such flows to high-resolution with stage \uppercase\expandafter{\romannumeral2}. As a result, our model can be effectively adapted to high-resolution videos.

\subsection{Loss function}
\label{sec:3-4}

Following RIFE, we adopt a self-supervised privileged distillation scheme to supervise the estimated flows directly. In detail, an additional teacher IFBlock is stacked to refine the estimated flows using the current frame $I_t$ as input, and the generated $F^{Tea}$ and $M^{Tea}$ can supervise the intermediate flows with a distillation loss:
\begin{equation}
    L_{dis} = \sum_{i\in\{0,1\}} ||F_{t\to i}-F_{t\to i}^{Tea}||
\end{equation}
which is applied over all estimated flows from each IFBlock. The gradient of the distillation loss is stopped for the teacher module.

The overall training loss consists of the reconstruction loss of the student $L_{rec}$, the teacher $L_{rec}^{Tea}$ and the privileged distillation loss $L_{dis}$:
\begin{equation}
    L = L_{rec}+L_{rec}^{Tea}+\lambda_d L_{dis}
\end{equation}
where the reconstruction loss is defined as the $L_1$ loss between the Laplacian pyramid representations of the synthesized frame and the ground truth. $\lambda_d$ is set to 0.01 by default.

\section{Experiments}
\label{sec:exps}

\begin{table*}
    \caption{Quantitative comparison(PSNR/SSIM) of video frame interpolation results. The best result in each set in shown in \textcolor{red}{red} and the second best is shown in \textcolor{blue}{blue}. Runtime and parameters are tested on the same device under the same settings, except for results with * copied from the original papers.}
    \label{tab1}
	\centering
    \resizebox{\linewidth}{!}{
	\begin{tabular}{c c c c c c c c c c c c}
	    \midrule
		\multirow{2}*{Model} & \multirow{2}*{X4K1000FPS}  &  \multicolumn{4}{c}{SNU-FILM}& \multirow{2}*{UCF101} & \multirow{2}*{Vimeo90K} & \multirow{2}*{\textbf{Average}} & \multicolumn{2}{c}{Runtime (ms)} & Parameters\\
		\cline{3-6} \cline{10-11}
		~ & ~  & Easy & Medium & Hard & Extreme & ~ & ~ & ~ & 480p & 1080p & (Million)\\
	    \midrule
		
		DAIN & 26.78/0.8065 & 39.73/0.9902 & 35.46/0.9780 & 30.17/0.9335 & 25.09/0.8584 & 34.99/0.9683 & 34.71/0.9756 & 32.42/0.9300 & 130* & - & 24.0* \\

		AdaCoF & 23.90/0.7271 & 39.80/0.9900 & 35.05/0.9754 & 29.46/0.9244 & 24.31/0.9439 & 34.90/0.9680 & 34.47/0.9730 & 31.70/0.9288 & 36 & 210 & 21.8 \\
		
		XVFI & 30.12/0.8704 & 39.92/0.9902 & 35.37/0.9777 & 29.57/0.9272 & 24.17/0.8448 & 35.18/0.9685 & 35.07/0.9756 & 32.77/0.9363 & 60 & 398 & 5.7  \\
		
		SoftSplat & -  & - & - & - & -& 35.39/\textcolor{red}{0.9700} & 36.10/0.9800 & - & 135* & - & 7.7* \\
		
		BMBC &29.35/0.8791& 39.90/0.9902 & 35.31/0.9774 & 29.33/0.9270 & 23.92/0.8432 & 35.15/0.9689 & 35.01/0.9764 & 32.57/0.9374 & 894 & 5787 & 11.0  \\
		
		ABME &  30.16/0.8793 & 39.59/0.9901 & 35.77/0.9789 & 30.58/\textcolor{blue}{0.9364} & 25.42/\textcolor{blue}{0.8639} & 35.38/0.9698 & \textcolor{blue}{36.18}/\textcolor{blue}{0.9805} & 
		33.30/0.9425 & 226 & 1386 & 17.6  \\
		
		RIFE & 29.14/0.8765& 40.02/0.9905  & 35.73/0.9787 & 30.08/0.9328 & 24.82/0.8530  & 35.28/0.9690 & 35.61/0.9779 & 32.95/0.9398 & 12 & 56 & 10.1 \\
		
		RIFE-Large & 28.94/0.8721 & \textcolor{red}{40.23}/\textcolor{red}{0.9907}  & 35.86/\textcolor{blue}{0.9792} & 30.19/0.9332 & 24.81/0.8540 & \textcolor{blue}{35.41}/\textcolor{red}{0.9700} & 36.13/0.9800 & 33.08/0.9399 & 54 & 350 & 10.1 \\
		
	    \midrule
		Ours-Base  &  \textcolor{blue}{31.63}/\textcolor{blue}{0.9185} &
		39.98/0.9903 & \textcolor{blue}{35.94}/0.9788 & \textcolor{blue}{30.72}/0.9359 & \textcolor{blue}{25.55}/0.8624& 35.36/0.9695& 35.88/0.9795 & 
		\textcolor{blue}{33.58}/\textcolor{blue}{0.9478} &
		38 & 82 & 12.1 \\
		
		Ours-Large  &  \textcolor{red}{31.86}/\textcolor{red}{0.9225} & \textcolor{blue}{40.14}/\textcolor{blue}{0.9905} & \textcolor{red}{36.12}/\textcolor{red}{0.9793} & \textcolor{red}{30.88}/\textcolor{red}{0.9370} & \textcolor{red}{25.70}/\textcolor{red}{0.8647} & \textcolor{red}{35.43}/\textcolor{red}{0.9700} & \textcolor{red}{36.22}/\textcolor{red}{0.9807} &
		\textcolor{red}{33.76}/\textcolor{red}{0.9492}  &
		122 & 419 & 12.1 \\
	    \midrule
		
	\end{tabular}
	}
\end{table*}

\begin{table*}
    \caption{Quantitative comparison(PSNR/SSIM) of video frame extrapolation results. The best result in each set in shown in \textcolor{red}{red} and the second best is shown in \textcolor{blue}{blue}.}
    \label{tab2}
	\centering
    \resizebox{\linewidth}{!}{
	\begin{tabular}{c c c c c c c c c c c c}
	    \midrule
		\multirow{2}*{Model} &  \multirow{2}*{X4K1000FPS} &  \multicolumn{4}{c}{SNU-FILM}&  \multirow{2}*{UCF101} & \multirow{2}*{Vimeo90K} & \multirow{2}*{\textbf{Average}} & \multicolumn{2}{c}{Runtime (ms)}  &Parameters\\
		\cline{3-6}
		~ & ~ & Easy & Medium & Hard & Extreme & ~ & ~ & ~ & 480p & 1080p & (Million)\\
	    \midrule
	    
		DVF & 19.18/0.6879 & 25.39/0.8728 & 23.30/0.8279 & 21.41/0.7798 & 19.49/0.7251 & 31.29/0.9433 & 26.21/0.8815 & 23.75/0.8169 & 40 & 256 & 3.8 \\
	    
		RIFE & 26.34/0.8456 & \textcolor{red}{36.76}/\textcolor{red}{0.9821} & \textcolor{red}{32.31}/\textcolor{red}{0.9566} & 27.10/0.8880 & 22.59/0.8061 & 31.69/0.9441 & 31.68/0.9580 & 29.78/0.9115 & 12 & 56 & 10.1 \\

		RIFE-Large & 21.85/0.8014 & 33.02/0.9501 & 29.83/0.9274 & 25.78/0.8666 & 22.10/0.7941 & \textcolor{blue}{31.78}/\textcolor{blue}{0.9447} & \textcolor{blue}{32.07}/\textcolor{blue}{0.9611} & 28.06/0.8922 & 54 & 350 & 10.1 \\
	    \midrule
	    
		Ours-Base & \textcolor{blue}{27.69}/\textcolor{blue}{0.8694}& \textcolor{blue}{36.46}/\textcolor{blue}{0.9814} & 
		32.21/0.9557 & 
		\textcolor{blue}{27.28}/\textcolor{blue}{0.8881} & \textcolor{blue}{22.84}/\textcolor{blue}{0.8090}  & 31.72/0.9444 & 31.84/0.9595 & 
		\textcolor{blue}{30.01}/\textcolor{blue}{0.9154} & 38 & 82& 12.1 \\
		
		Ours-Large & \textcolor{red}{28.05}/\textcolor{red}{0.8786}& 36.40/\textcolor{blue}{0.9814} & 
		\textcolor{blue}{32.24}/\textcolor{blue}{0.9561} & 
		\textcolor{red}{27.41}/\textcolor{red}{0.8895} & \textcolor{red}{23.00}/\textcolor{red}{0.8120}  & \textcolor{red}{31.81}/\textcolor{red}{0.9449} & \textcolor{red}{32.10}/\textcolor{red}{0.9614} & 
		\textcolor{red}{30.14}/\textcolor{red}{0.9177} & 122 & 419& 12.1 \\
	    \midrule
		
	\end{tabular}
	}
\end{table*}

\subsection{Experimental Setup}
\label{sec:4-1}

\ \ \ \ \ \textbf{Training Data}. We use the Vimeo-90k \cite{xue2019video} training split, which has 51,312 triplets with a resolution of $448\times256$. We augment the dataset by randomly flipping, temporal order revering, and cropping $256\times256$ patches.

\textbf{Training Strategy}. We use AdamW \cite{loshchilov2018fixing} to optimize our model with weight decay $10^{-3}$. The learning rate is gradually reduced from $3\times10^{-4}$ to $3\times10^{-5}$ using cosine annealing for each stage in progressive learning. The batch size is set to 64, and we use 4 Telsa V100 GPU to train the coarse-scale module for 230k iterations in stage \uppercase\expandafter{\romannumeral1} and train the other parts for 76k iterations in stage \uppercase\expandafter{\romannumeral2}. It takes about 40 hours for training in total.

\subsection{Benchmarks}
\label{sec:4-2}

We evaluate our scheme on various benchmarks to verify its performance and generalization ability. On each dataset, we measure the peak signal-to-noise ratio (PSNR) and structural similarity (SSIM) for quantitative evaluation.

\textbf{X4K1000FPS} \cite{sim2021xvfi}: It is a high-quality dataset of 4K videos, which is challenging due to the high resolution, occlusion, large motion and the scene diversity. The provided X-TEST set supports $8\times$ interpolation evaluation on two frames with a temporal distance of 32 frames, and we also use it to evaluate $2\times$ extrapolation by synthesising the $32-$nd frame using the $0-$th and the $16$-th. 

\textbf{SNU-FILM} \cite{choi2020channel}: It contains 1,240 triplets of 240fps videos with resolution from $640\times368$ to $1280\times720$. Four settings - Easy, Medium, Hard, and Extreme - are provided to evaluate from small motion to large motion, and the temporal distance of each setting increases from 2 (120fps $\to$ 240fps) to 16 (15fps $\to$ 30fps). 

\textbf{UCF101} \cite{soomro2012ucf101}: Liu et al. \cite{liu2017video} selected 379 triplets from UCF101 human actions dataset for video frame synthesis evaluation. However, there are some dirty data in the test set. We will demonstrate it in detail in the supplementary material.


\textbf{Vimeo90K} \cite{xue2019video}: The test set of Vimeo90K contains 3,782 triplets with resolution of $448\times256$. We evaluate on it to verify the robustness of our model on low-resolution videos.

Different downsample size $(h,w)$ is set to adapt to the resolution of different benchmark. We set $a=384$ for X4K1000FPS and $a=256$ for other benchmarks (the definition of $a$ is in Section \ref{sec:3-3}).

\begin{table*}[h]
    \caption{Add on study of progressive learning and NCM. PSNR/SSIM and the runtime(ms) are reported.}
    \label{tab3}
	\centering
    \resizebox{0.85\linewidth}{!}{
	\begin{tabular}{c c c c c c c c}
	    \midrule
		Progressive & \multicolumn{2}{c}{NCM} &  \multirow{2}*{X4K1000FPS}     & \multicolumn{2}{c}{Vimeo90K} & Runtime \\
		\cline{2-3}  \cline{5-6}
		Learning & image feature & motion feature & ~ & 256p & 480p & @1080p \\
	    \midrule
	    \checkmark  & \checkmark & \checkmark & \textbf{31.63}/\textbf{0.9185} &  \textbf{35.88}/\textbf{0.9795} & \textbf{36.44}/\textbf{0.9816} & 82 \\
	    \checkmark & \checkmark & $\times$ & 30.98/0.9087 & 35.76/0.9788 & 36.35/0.9812 & 69\\
	    \checkmark  & $\times$ & $\times$ & 30.93/0.9088 & 35.36/0.9773 & 36.07/0.9803 & \textbf{63}\\
	    $\times$ & $\times$ & $\times$ & 30.00/0.8937 & 35.53/0.9779 & 35.93/0.9796 & \textbf{63} \\
	    \midrule
	\end{tabular}
	}
\end{table*}

\begin{table}[t]
    \caption{Ablation study on coarse-to-fine manner and matching region. PSNR/SSIM and runtime(ms) are reported. }
    \label{tab4}
	\centering
    \resizebox{\linewidth}{!}{
	\begin{tabular}{l c c c c c}
	    \midrule
		   \multirow{2}*{Setting}& \multirow{2}*{X4K1000FPS} & \multirow{2}*{Vimeo90K} & Runtime \\
		   ~ & ~ & ~ & @1080p \\
	    \midrule
		w/ normal coarse-to-fine & 31.35/0.9151 & \textbf{35.93}/\textbf{0.9797} & 244\\
	    w/ flow-guided matching & 31.56/0.9180 & 35.83/0.9793 & 92\\
	    \midrule
		Ours-base & \textbf{31.63}/\textbf{0.9185} & 35.88/0.9795 & \textbf{82}  \\
		\multicolumn{4}{l}{(Heterogeneous coarse-to-fine, neighbor matching)}  \\
	    \midrule
	\end{tabular}
	}
\end{table}

\subsection{Comparison with Previous Methods}
\label{sec:4-3}

For video frame interpolation, we compare the proposed scheme with previous methods : DAIN \cite{bao2019depth}, AdaCoF \cite{lee2020adacof}, XVFI \cite{sim2021xvfi}, SoftSplat \cite{niklaus2020softmax}, BMBC \cite{park2020bmbc}, ABME \cite{park2021asymmetric} and RIFE \cite{huang2020rife}. For extrapolation, we compare with DVF \cite{liu2017video}, and re-implement RIFE on the extrapolation task for comparison. 

\begin{figure}[h]
    \includegraphics[width=\linewidth]{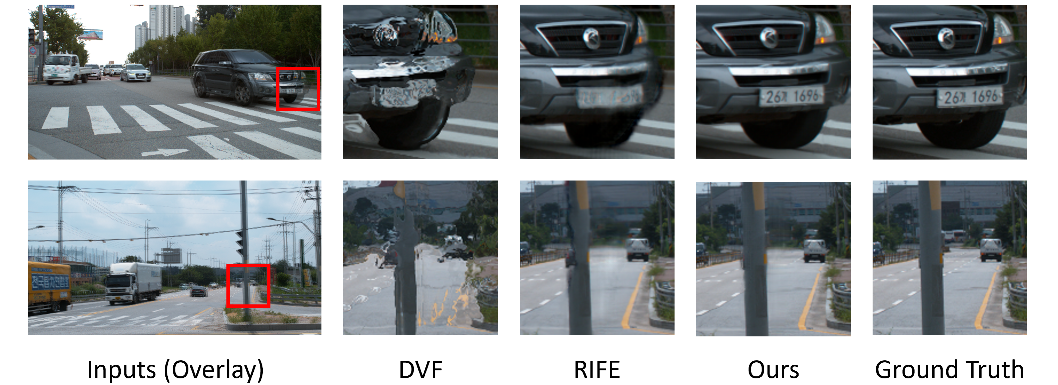}
    \caption{Visual comparisons on extrapolation with DVF \cite{liu2017video} and RIFE \cite{huang2020rife}. \textit{Best viewed in zoom.}}
    \label{fig6}
\end{figure}

Following RIFE \cite{huang2020rife}, we also introduce a \textit{Large} version of our model to meet the need of different scenarios with different computation cost. Two modification are performed : 1) \textit{test-time augmentation}, to inference twice with the original input frames and the flipped frames then average the results, and 2) \textit{model scaling}, to double the resolution of feature map in each IFBlock and the synthesis network by removing the first stride. More information can be found in the supplementary material.

We report the quantitative results of interpolation in Table.\ref{tab1}. Our large model achieves the best performance in all benchmarks except for the SNU-FILM Easy setting. Compared with previous state-of-the-art scheme ABME, our base model improves the average PSNR by 0.27dB with 17 times faster runtime in 1080p videos, and our large model improves by 0.46dB with 3.3 times faster runtime. It is worth noting that our base and large models outperform ABME by surprising 1.47dB (30.16dB$\to$31.63dB) and 1.70dB (30.16dB$\to$31.86dB) in the most challenging X4K1000FPS benchmark, which shows its capability in capturing large motion in high-resolution videos. Compared with the real-time scheme RIFE-Large, our base model shows 0.50 dB improvement with 4.2 times faster runtime. Visual comparison can be found in Fig.\ref{fig1} and the supplementary material.

We also report the quantitative results of extrapolation in Table.\ref{tab2}, and the visual comparison is shown in Fig.\ref{fig6}. Compared with RIFE, our model achieves better performance in all benchmarks except for SNU-FILM Easy and Medium setting, leading to an average improvement of 0.23dB (base model) and 2.08dB (large model) in terms of PSNR. And we find that RIFE-Large shows much worse performance than RIFE in high-resolution videos (e.g., 4.49dB on X4K1000FPS and 2.02dB drop on SNU-FILM), which is also observed on the interpolation benchmark where RIFE-Large shows 0.2dB drop in X4K1000FPS. It is because when doubling feature resolution, the receptive field of the network is halved. However, our model is not affected much by the resolution of feature map because the receptive field is guaranteed by NCM instead of convolution layers. It means our scheme can be better extended to the large version for better performance.

\subsection{Ablation Study}

\subsubsection{Add-on Study.}
To understand the effectiveness of the proposed components, we perform an ablation study to add each component on a baseline in Table\ref{tab3}. The baseline (last line in the table) consists of five IFBlocks, and between each two blocks the upsampling rate is set to 2 in both training and inference. 

\textbf{Progressive learning}. It makes the model adapt to high resolution 4K videos (0.93 dB improvement on X4K1000FPS). However, the performance on 256p videos sightly drops. This is because the baseline is only trained for low-resolution 256p videos, while progressive learning is designed to refine high-resolution videos on low-resolution 256p videos. So if we resize 256p videos to 480p, such limitation is eliminated and achieves 0.14dB improvement.

\textbf{Neighbor correspondence matching} comprises of a feature pyramid to extract \textit{image features} and a matching process to extract \textit{motion features}. The image features mainly improves performance on low-resolution videos, and the matching-based motion features can enhance performance on both high-resolution (0.65dB) and low-resolution (0.12dB).

\subsubsection{Coarse-to-fine and the matching region.}
We also perform an ablation study on the coarse-to-fine structure and the matching region in NCM in Table.\ref{tab4}. 

\textbf{Heterogeneous coarse-to-fine}. Compared with using NCM in both coarse and fine-scale (normal coarse-to-fine in Table.\ref{tab4}), the proposed heterogeneous coarse-to-fine causes 3 times shorter runtime with better performance on 4K videos and comparable performance on 256p videos. It shows the efficiency to design different module structure for the coarse and fine-scale modules. Using NCM in the fine-scale module in 4K videos can cause performance drop, because in such high-resolution videos, the content is so similar in the neighbor regions that the correspondence is not effective to capture the motion.

\textbf{Matching region}. We compare the proposed neighbor correspondence matching with the flow-guided matching. Guiding matching by flow causes performance to drop by 0.07dB in X4K1000FPS. It is because the matching region may be misled by the inaccurate estimated flows. And runtime is also longer by 10ms since more matching operations are needed. In addition, we find that using flow-guided matching is more likely to cause collapse in training. It indicates the flow-guided matching is less stable since the matching region is influenced by estimated flows.

\section{Application : Video Compression}

The proposed scheme can be applied to various scenarios due to the powerful capability in capture large motion in high-resolution videos. For example, it can be applied in video compression, jitters removal in real-time communication system, and motion effects generation. In this section, we use video compression as an example to show its potential. Experiments and demos about other applications an be found in the supplementary material. 

Most video compression schemes  \cite{lu2019dvc,lu2020end,yang2020learning,li2021deep,sheng2021temporal} adopt a motion estimation and motion compensation (MEMC) paradigm. They predict the current frame using the optical flow, and then compress the prediction residual to reconstruct the compressed frame. Many works \cite{yang2020learning,li2021deep,sheng2021temporal} focus on improving the residual compression process, but the temporal redundancy in the optical flow is not well considered. We can eliminate such redundancy partly by using the proposed module for motion prediction.We design a scalable bi-directional video compression model using NCM. It can serve as a plugin-in on any uni-direction codec (e.g.,  P-frame codec  \cite{sheng2021temporal, HM}) to compress B-frame, and the whole video can be compressed in order of I-B-P-B-$\cdots$. The model supports a \textit{bit-free} mode for low-cost compression and a \textit{bit-need} mode for high-quality compression.

\begin{figure}[t]
    \includegraphics[width=\linewidth]{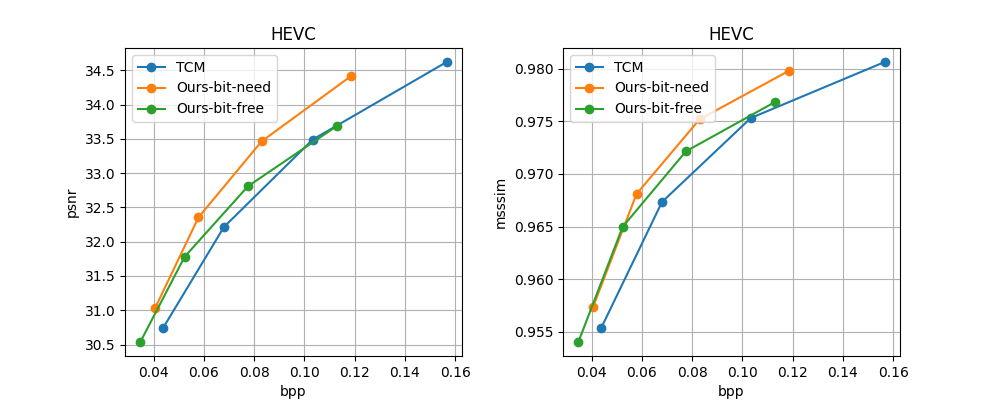}
    \caption{Apply the proposed frame synthesis model in video compression. The designed plugin-and-in bi-directional codec can achieve higher compression rate compared with the baseline TCM  \cite{sheng2021temporal}. }
    \label{fig8}
\end{figure}

The bit-free mode simply adopts our video frame synthesis model to interpolate the B-frame. It means the codec does not need to encode and decode code streams of B-frame, and this frame can be interpolated using much less cost. We evaluate it on the state-of-the-art learned P-frame codec TCM  \cite{sheng2021temporal}, and the rate-distortion curve on HEVC test videos \cite{sullivan2012overview} is shown in Fig.\ref{fig8}. Our model can save 10.4\% bits (0.065bpp$\to$0.058bpp) under PSNR=32.0dB and save 17.2\% bits (0.063bpp$\to$0.052bpp) under MS-SSIM=0.9650. In addition, our model can inference about 9 times faster in B-frames, from 500ms encoding and 252ms decoding time of TCM to 0ms encoding and 82ms interpolation time. It demonstrates the efficiency of bit-free mode in low-cost scenarios such as video communication.

\begin{figure}[t]
    \includegraphics[width=\linewidth]{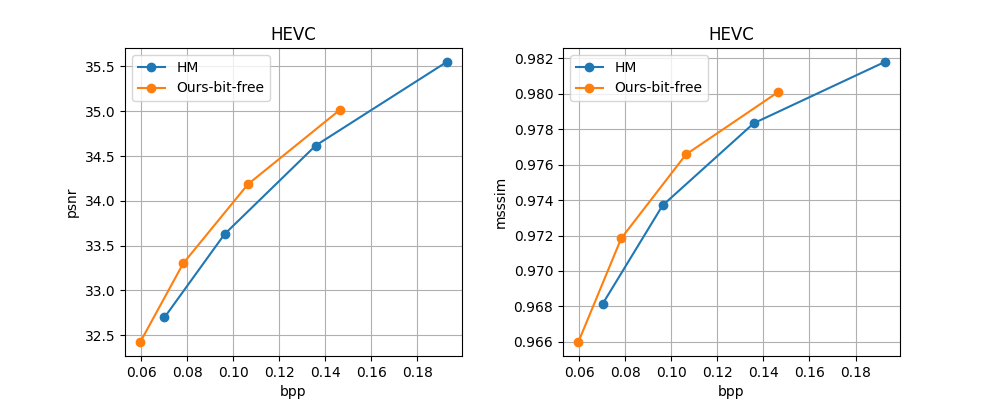}
    \caption{Compare the proposed B-frame video codec with H.265-HM \cite{HM}.}
    \label{fig9}
\end{figure}

The bit-need mode adopts our video frame synthesis model as a motion prediction module, and compress the residual motion instead of the entire motion. Experiments show it can save 17.5\% bits (from 0.065bpp to 0.053bpp) under PSNR=32.0 on TCM. We also compare it with H.265-HM \cite{HM} under the same sequence order in Fig.\ref{fig9}, and results show 10\% bits saving (from 0.111bpp to 0.100bpp) on the whole sequence under PSNR=34.0dB. It shows our model can serve as a superior motion prediction module for video compression. More details of experiment settings and experiment results can be found in the supplementary material.

\section{Limitations}

\begin{figure}[t]
    \includegraphics[width=\linewidth]{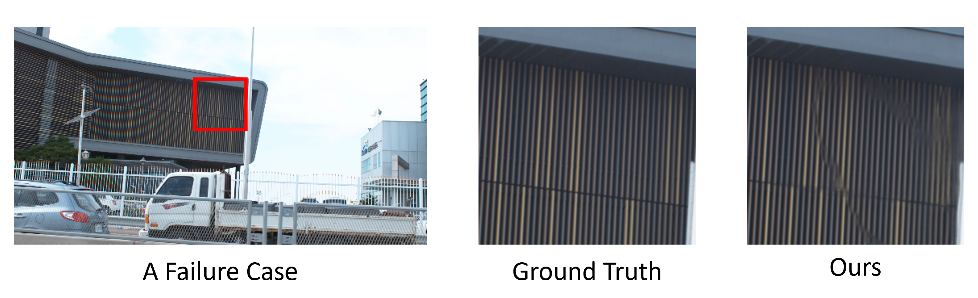}
    \caption{A failure case. Our model cannot estimate accurate flows on large region of repeated stripes, which may cause artifacts in the synthesised frame.}
    \label{fig7}
\end{figure}

Even the proposed neighbor correspondence matching algorithm can capture large motion for high-resolution videos, it may estimate inaccurate flows on similar details in the video. Because the matching is guided by feature similarity, neighbor regions with similar appearance cannot be distinguished by matching. As shown in Fig.\ref{fig7}, our model fails on estimating repeated stripes on the house. It also limits NCM to be used in fine-scale module in 4K videos because in such scale the neighbor region is usually full of similar regions. We hope to solve it in the future by distinguishing regions with similar appearance using techniques like positional embedding.

\section{Conclusion}

In this paper, we propose a neighbor correspondence matching algorithm for video frame synthesis, which can capture large motion even in high-resolution videos. With the proposed heterogeneous coarse-to-fine structure design and the progressive learning, our model is both effective and efficient and can adapt to high-resolution videos. Experiments show the superiority of our model in both interpolation and extrapolation. In addition, our model can be used in many applications, and we use video compression as an example to show its potential. We hope it can be extended to more applications in the future.

\newpage

\bibliographystyle{ACM-Reference-Format}
\balance
\bibliography{sample-base}

\newpage

\appendix

\section{Details of Experiments}

\subsection{Test-Time Augmentation and Model Scaling}
Following RIFE\cite{huang2020rife}, we also introduce a \textit{Large} version of our model by performing test-time augmentation and model scaling. We use the same strategy as in RIFE : 1) We flip the input frames horizontally and vertically as the augmented test data, and infer the results on the augmented data. The results are flipped back and then averaged to generate the final results. 2) We double the resolution of the feature map. For each IFBlocks and the synthesis network, we remove the stride on the first convolution layer and the last transposed convolution layer.

\subsection{Runtime Analysis}

\begin{figure}[h]
    \centering
    \includegraphics[width=0.85\linewidth]{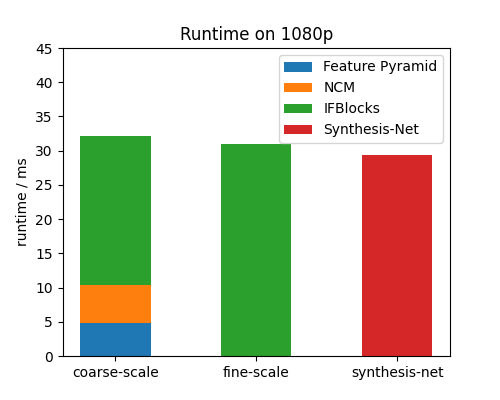}
    \caption{Runtime of each components.}
    \label{fig-sm1}
\end{figure}

We also visualize the runtime of different components in Fig.\ref{fig-sm1}. Since the feature pyramid and the neighbor correspondence matching are only performed in coarse-scale, it will not cause much time cost. The main time cost comes from stacked convolutional layers in IFBlocks and the synthesis network.

\subsection{Extrapolation Settings}
The extrapolation task is more difficult to interpolation, so we adopt different experimental settings. First, We remove the distillation loss because we find that it leads to collapse in training. Second, instead of directly estimate the intermediate flow $(F_{t\to0},\ F_{t\to1})$, we estimate $(dF_{01}, F_{t\to1})$ and compute $F_{t\to0}=2\times F_{t\to1}+dF_{01}$. It avoids the unbalance temporal distance between current frame and two reference frames.

\subsection{Baseline Settings}
All baselines are tested using the available official weights except for RIFE for extrapolation, because RIFE is not implemented for extrapolation in the original paper. We re-implement RIFE as the baseline for extrapolation. We find setting base learning rate to $3\times10^{-4}$ cases collapse in training, so we set it to $1\times10^{-4}$. The distillation loss is not used since it also causes collapse.

In RIFE, we can change the downsample rate $K$ to adapt to different resolutions. It is set to $(4,2,1)$ in SNU-FILM, UCF101 and Vimeo90K benchmark like the official recommendation, and is set to $(16,4,1)$ in X4K1000FPS to enlarge receptive fields. We find it to perform much better than $(4,2,1)$ in X4K1000FPS (from PSNR=24.57dB to 28.94dB in interpolation).

\subsection{Dirty Data in UCF101}

\begin{figure}[h]
    \centering
    \includegraphics[width=\linewidth]{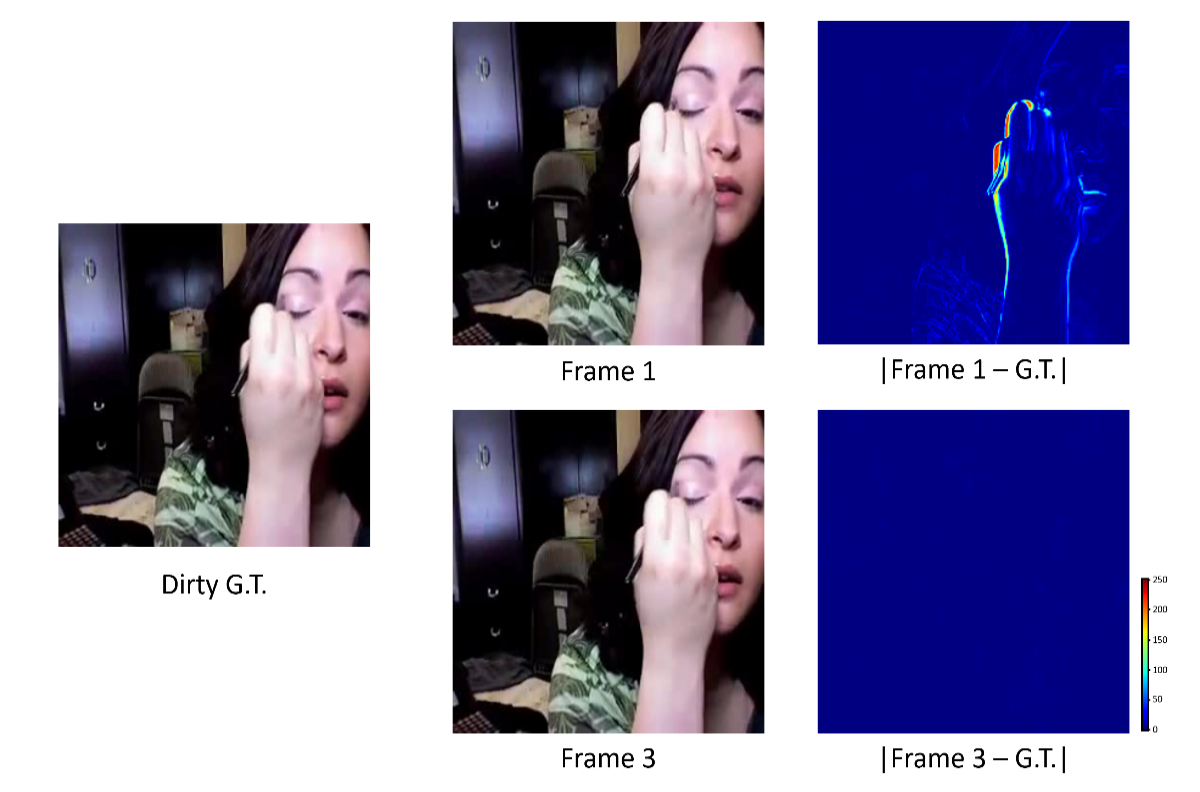}
    \caption{An example of dirty data in UCF101.}
    \label{fig-sm2}
\end{figure}

Liu et al.\cite{liu2017video} selected 378 triplets from UCF101 as test set for video frame synthesis. However, we find some dirty data (98 triplets) in the test set. As shown in Fig.\ref{fig-sm2}, such dirty data have two same frames in the triplet (e.g., the 2nd G.T. and 3d frame are the same), which leads to wrong ground truth for evaluation.
We exclude these dirty data and perform comparison experiments with RIFE in Table.\ref{tab-sm1}. We display ID of these dirty data in Table.\ref{tab-sm2} for reproducibility.

\section{Visual Comparison}

We conduct the visual comparison with XVFI\cite{sim2021xvfi} and ABME\cite{park2021asymmetric} for interpolation, and with RIFE and DVF\cite{liu2017video} for extrapolation. Compared with previous methods, our network can capture larger motion and more details. In addition, our network can preserve more complete structural information.

\begin{table}[h]
    \caption{Quantitative comparison(PSNR/SSIM) on UCF101. Dirty data are excluded. }
    \label{tab-sm1}
	\centering
	\begin{tabular}{c c c}
	    \midrule
	     Model  & Interpolation & Extrapolation \\
	    \midrule
		RIFE & 35.56/0.9745 & 31.95/0.9536 \\
		RIFE-Large & 35.67/0.9752 & 32.06/0.9545 \\
	    \midrule
		Ours-Base & 35.64/0.9752 & 31.98/0.9541 \\
		Ours-Large & \textbf{35.75}/\textbf{0.9757} & \textbf{32.08}/\textbf{0.9548} \\
	    \midrule
	\end{tabular}
\end{table}

\begin{table}[h]
    \caption{Dirty data IDs in UCF101.}
    \label{tab-sm2}
	\centering
	\begin{tabular}{c c c c c c c c c c}
	    \midrule
	     1& 141& 171& 191& 231& 241& 281& 291& 371\\
	     401& 441& 451& 511& 571& 611& 661& 711& 721\\
	     731& 971& 981& 1001& 1021& 1041& 1081& 1091& 1111\\
	     1221& 1231& 1241& 1251& 1261& 1321& 1361& 1401& 1411\\
	     1441& 1451& 1471& 1501& 1511& 1531& 1551& 1601& 1721\\
	     1741& 1811& 1901& 1911& 1961& 2051& 2061& 2111& 2191\\
	     2281& 2291& 2301& 2361& 2381& 2401& 2431& 2441& 2451\\
	     2471& 2481& 2491& 2531& 2671& 2691& 2711& 2751& 2761\\
	     2781& 2881& 2921& 2971& 3011& 3061& 3071& 3111& 3121\\ 3131& 3171& 3181& 3211& 3231& 3241& 3311& 3321& 3341\\
	     3361& 3381& 3391& 3441& 3601& 3651& 3661& 3721& \\
	    \midrule
	\end{tabular}
\end{table}

\begin{figure*}[h]
    \centering
    \includegraphics[width=\linewidth]{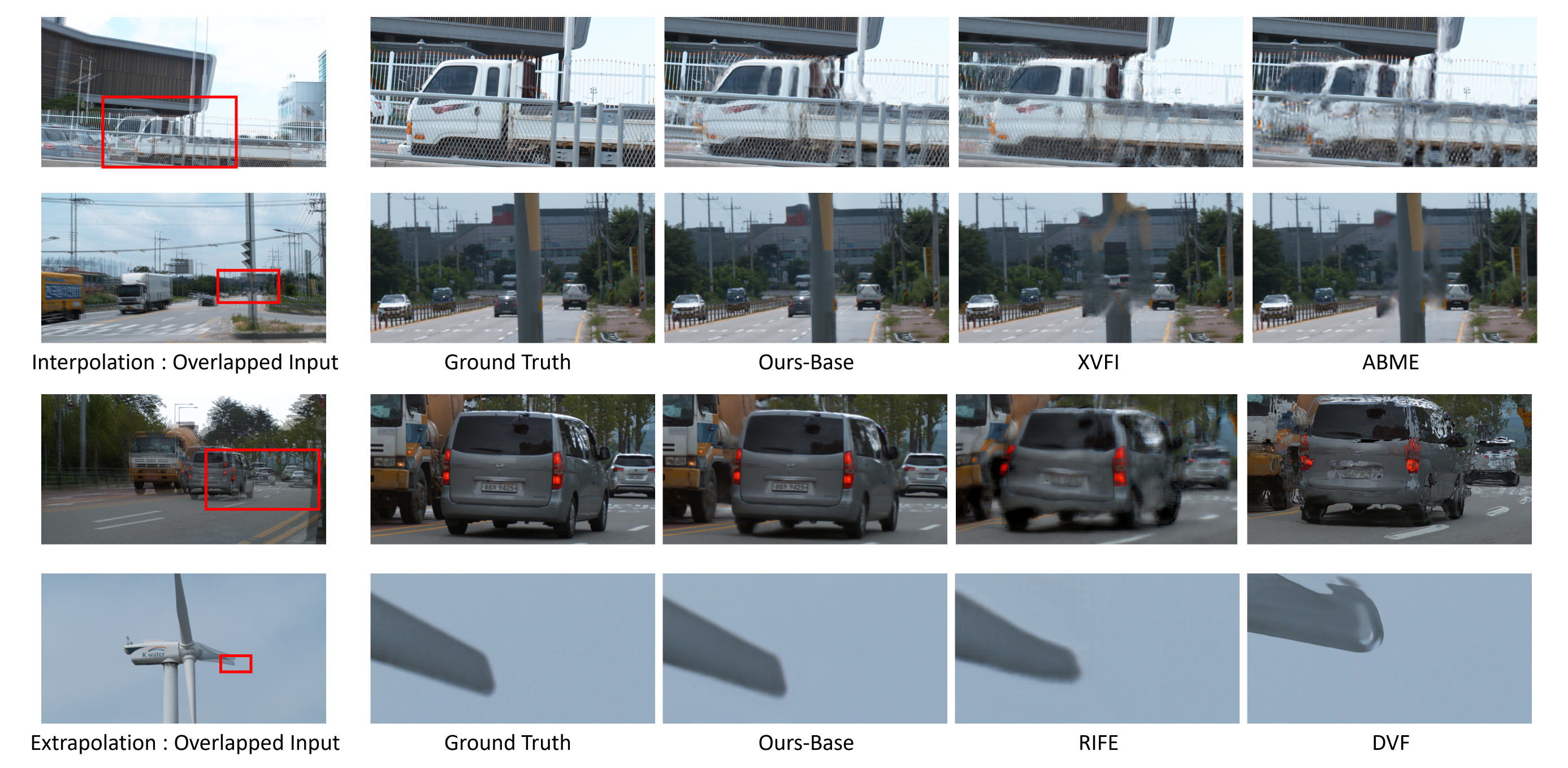}
    \caption{Visual quality comparison. Best view in zoom-in.}
    \label{fig-sm7}
\end{figure*}

\begin{figure*}[t]
    \centering
    \includegraphics[width=0.80\linewidth]{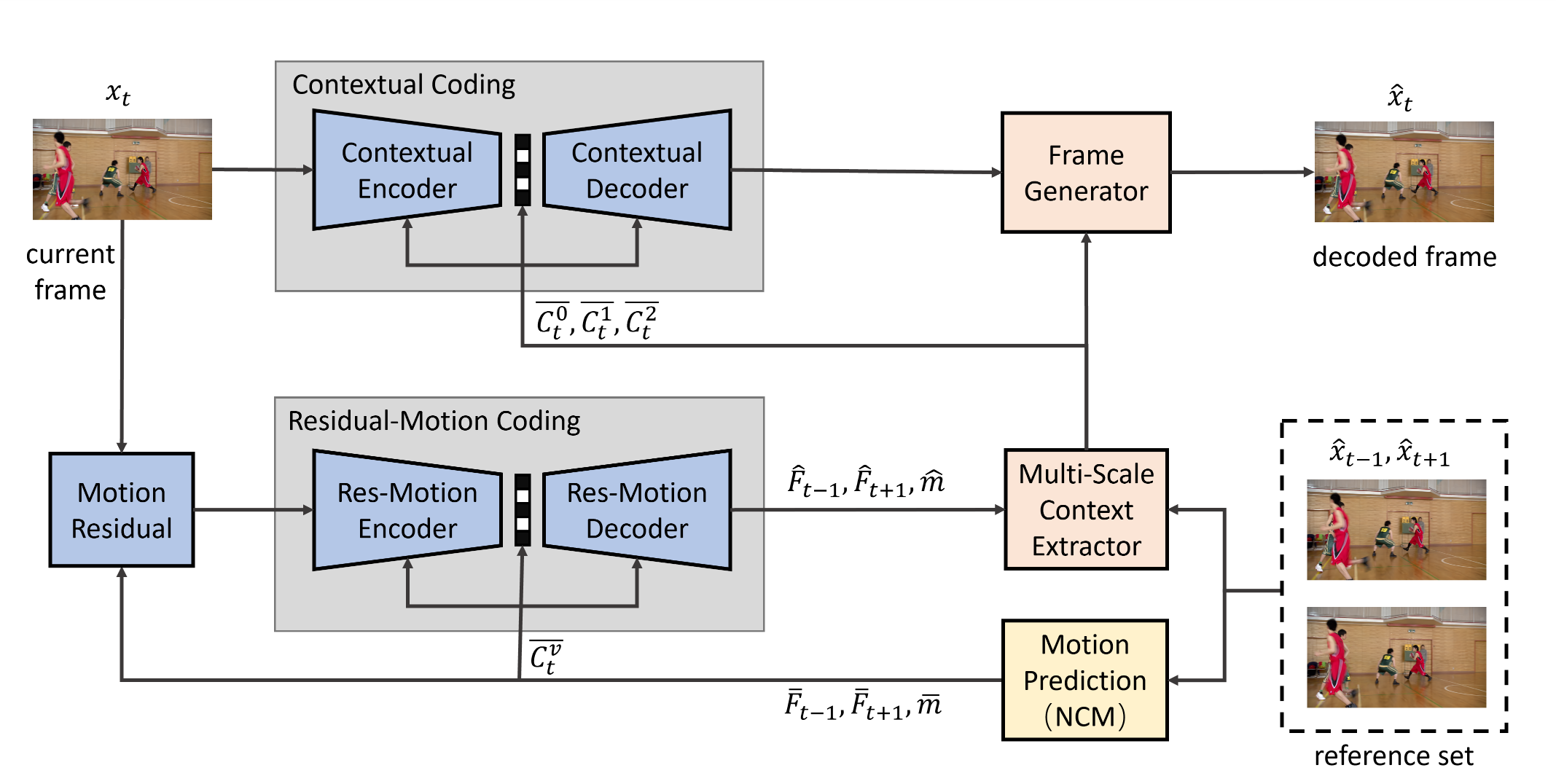}
    \caption{Model structure of the proposed bit-need bi-direction video codec.}
    \label{fig-sm3}
\end{figure*}

\begin{figure}[h]
    \centering
    \includegraphics[width=\linewidth]{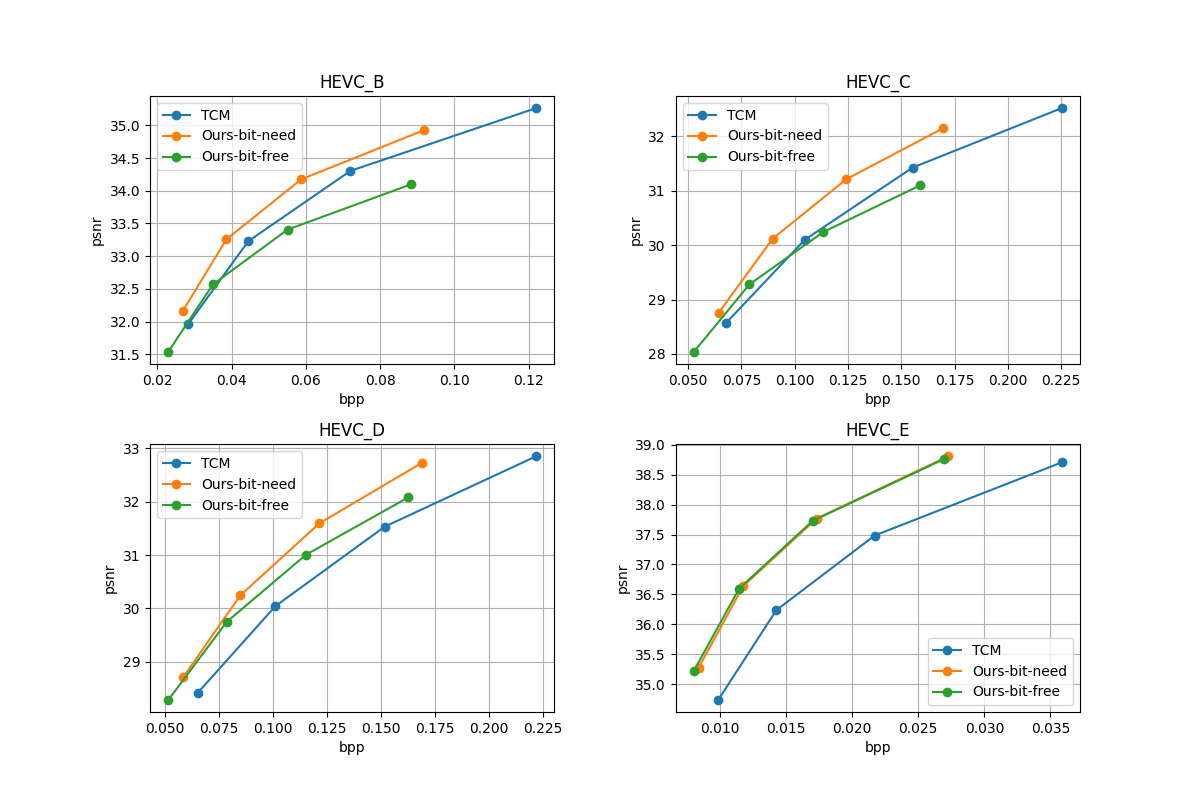}\\
    \includegraphics[width=\linewidth]{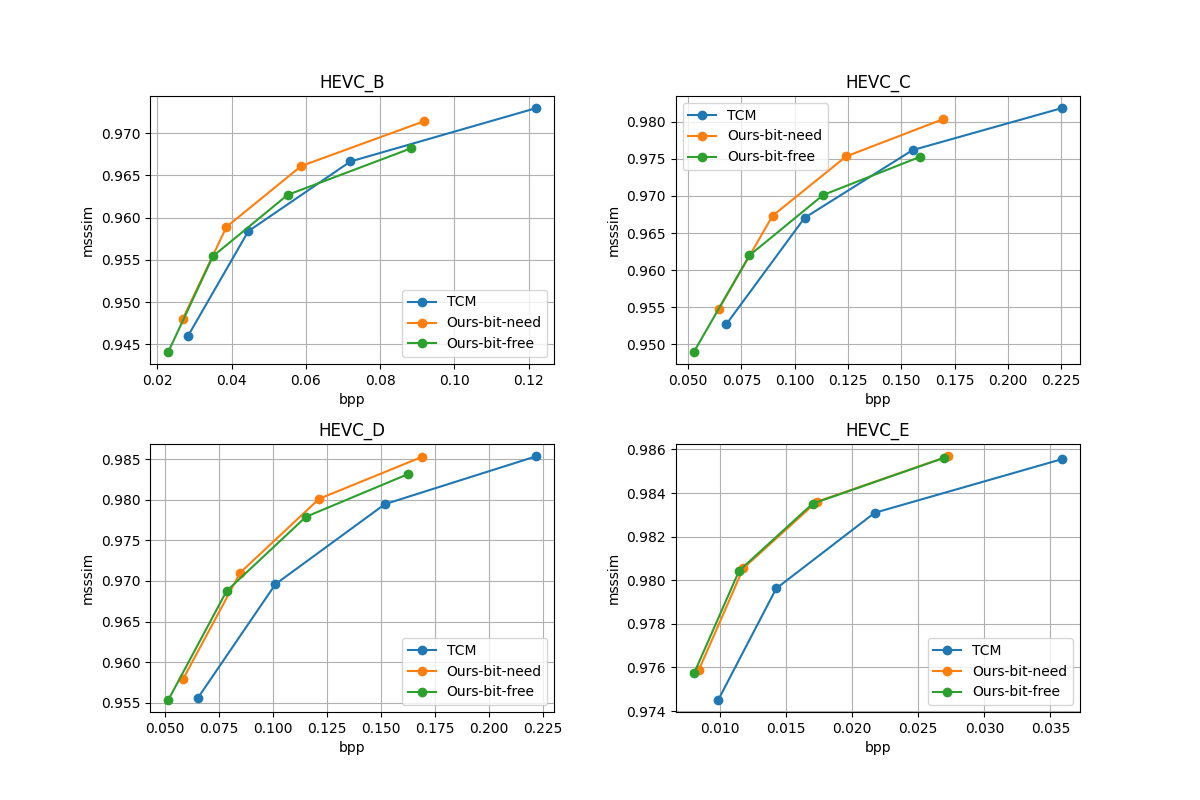}
    \caption{Compression experiment results on TCM.}
    \label{fig-sm4}
\end{figure}

\begin{figure}[h]
    \centering
    \includegraphics[width=\linewidth]{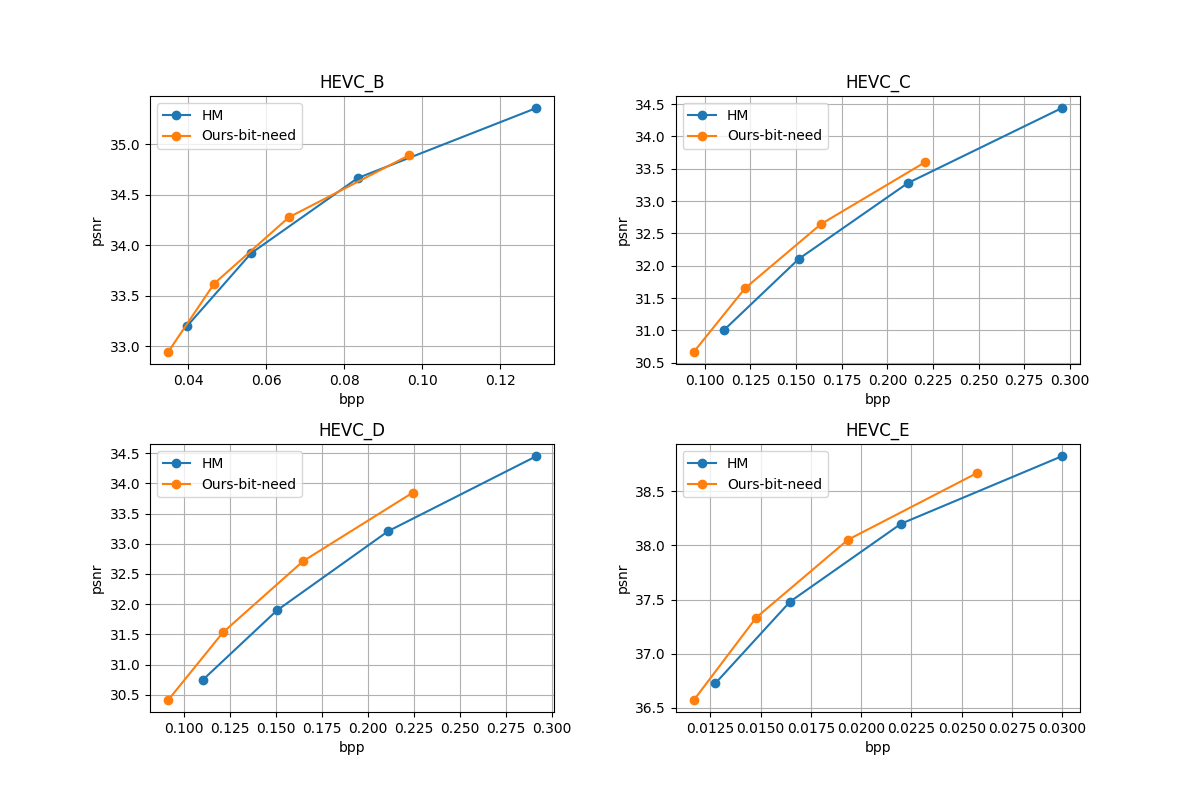}\\
    \includegraphics[width=\linewidth]{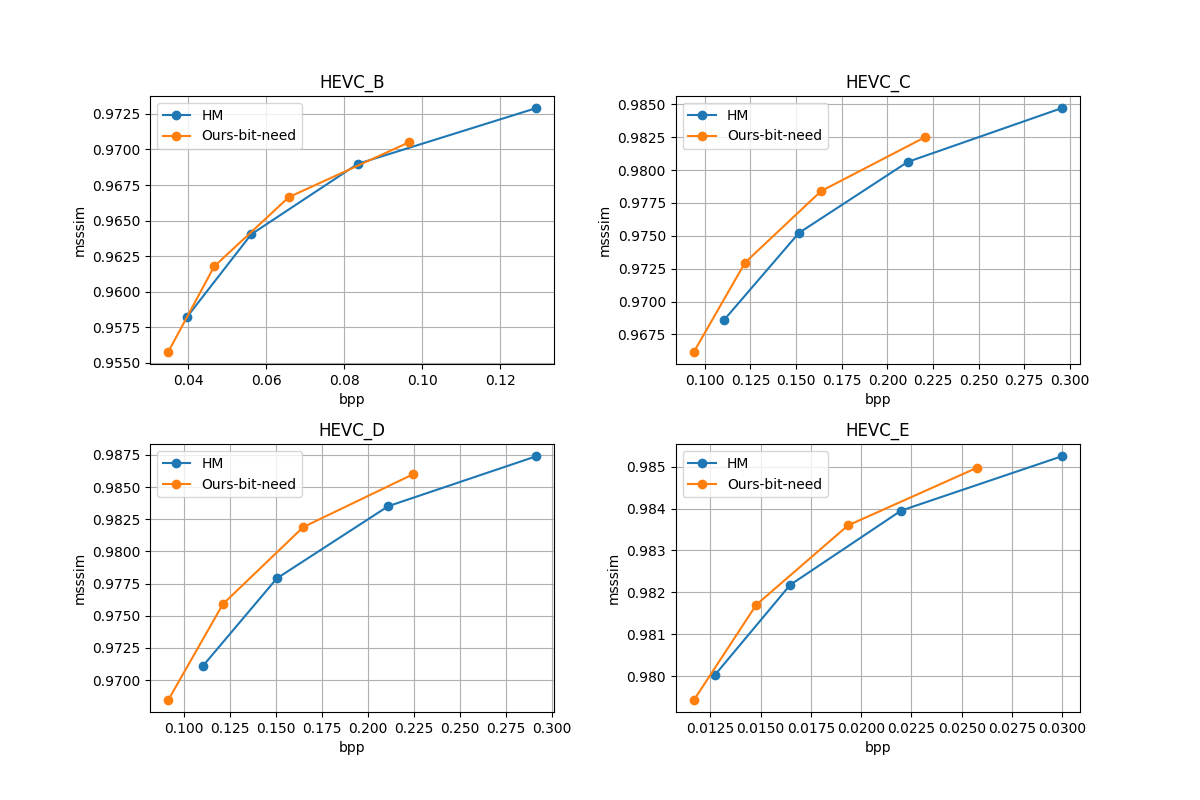}
    \caption{Compression experiment results on HM.}
    \label{fig-sm5}
\end{figure}

\begin{figure}[h]
    \centering
    \includegraphics[width=\linewidth]{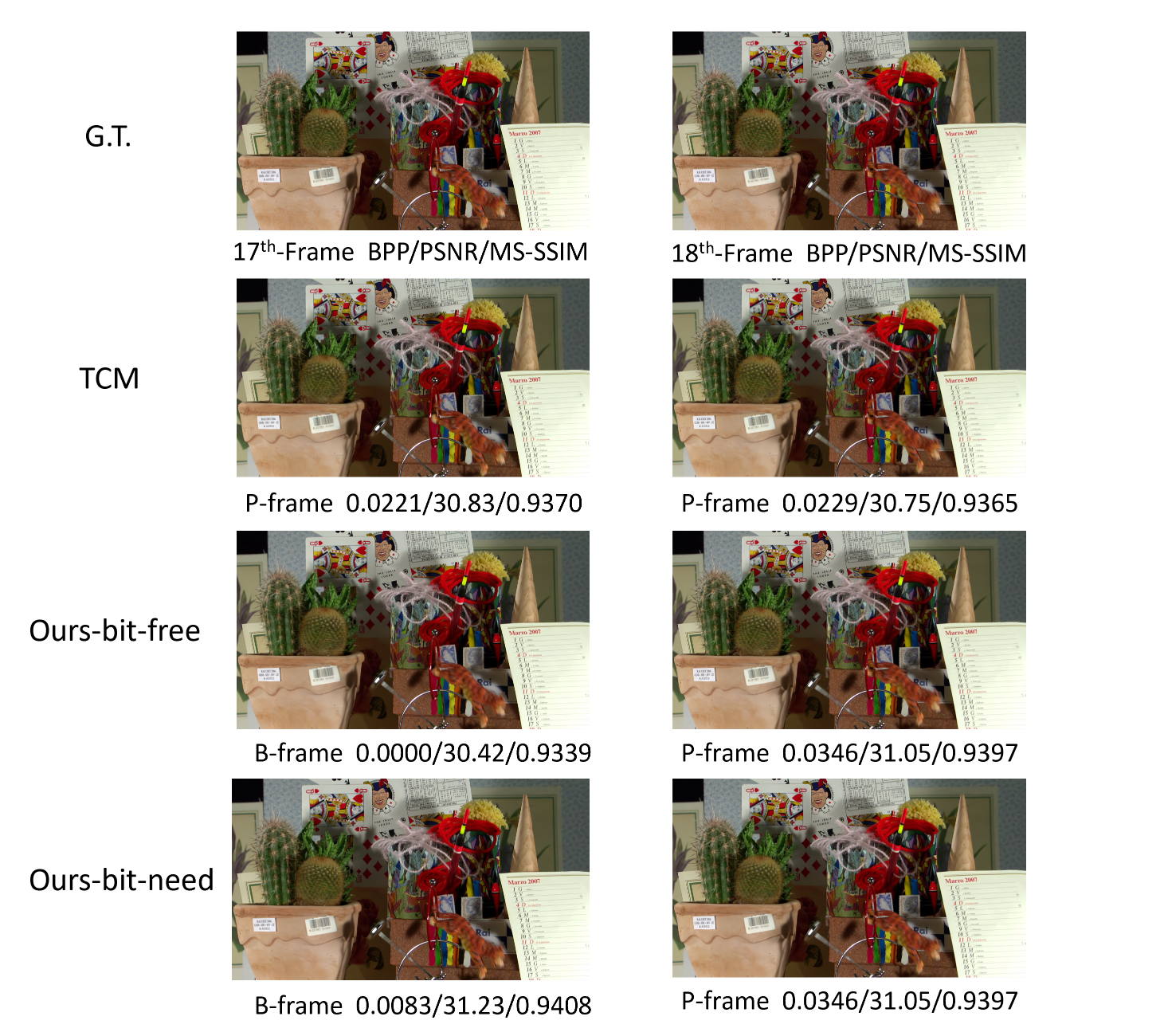}
    \caption{An exmaple of proposed video compression. Best view in zoom in.}
    \label{fig-sm8}
\end{figure}

\begin{figure}[h]
    \centering
    \includegraphics[width=\linewidth]{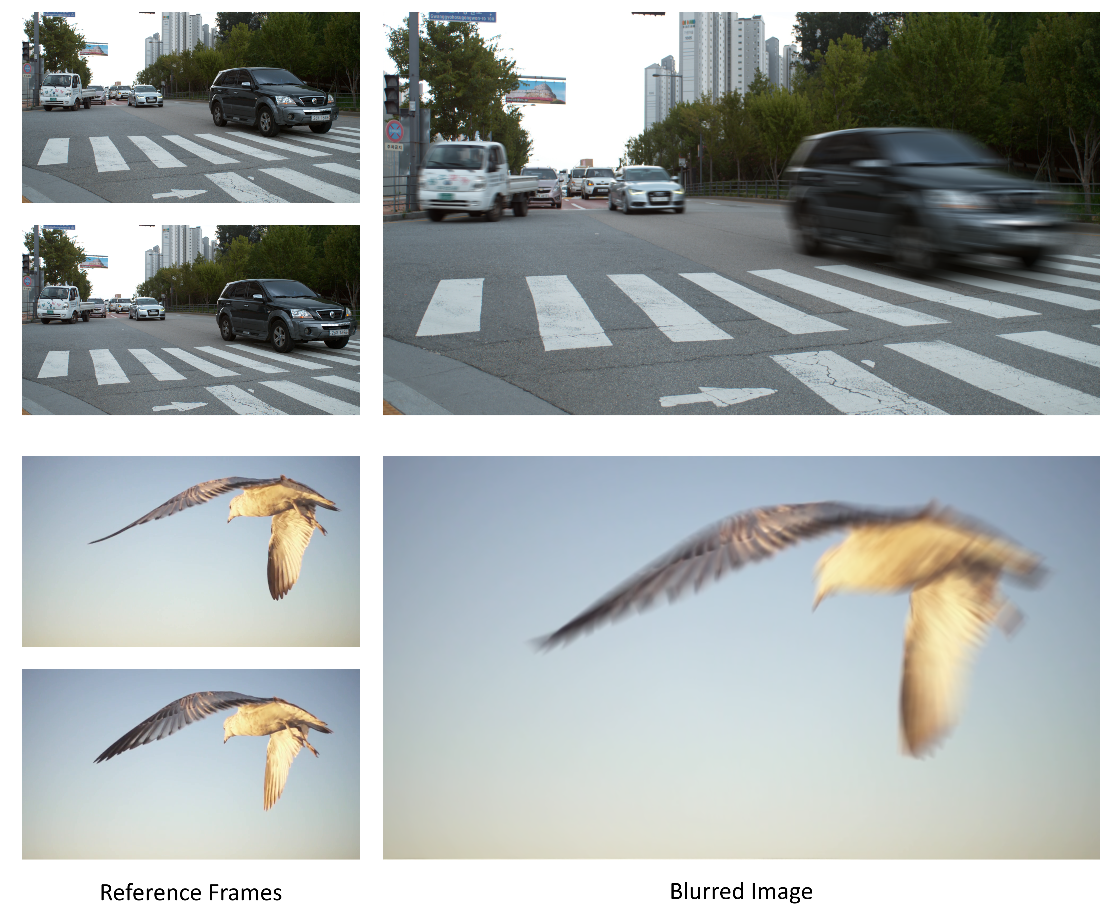}
    \caption{Motion blur generation.}
    \label{fig-sm6}
\end{figure}

\section{Application :  Video Compression}

In the paper, we apply the proposed frame synthesis network in video compression, and propose a plugin-and-in bi-directional video codec. It has two modes : a bit-free mode for low-cost and low-latency scenarios like video communication, and a bit-need mode for high-quality compression.

\subsection{Model Structure}

The bit-free mode adopts the proposed frame synthesis network to interpolate the B-frame directly. The bit-need mode consists of three parts as shown in Fig.\ref{fig-sm3} : 

\subsubsection{Motion Prediction.} It use the intermediate flow estimation part of proposed frame synthesis network to predict the intermediate flow and mask $(\Bar{F}, \Bar{M})$. The predicted motion can be obtained in both encoder and decoder without bits transmission.

\subsubsection{Motion Residual Refine.} When we train the frame synthesis network, a \textit{Teacher} block is also trained for the distillation loss. We directly adopt this block to refine the predicted flows with the current frame as input.

\subsubsection{Residual-Motion Coding.} We compress the residual motion between the refined flows and the predicted flows. We use a convolution layer to learn a motion context $\Bar{C_t^v}$ from the predicted flows, and perform contextual coding \cite{li2021deep} to compression the residual flow. Using the compressed residual motion, more accurate motion $\hat{F}, \hat{M}$ can be reconstructed in the decoder.

\subsubsection{Contextual Coding.} Following TCM \cite{sheng2021temporal}, we perform multi-scale context extracting using the compressed flows. The extracted contexts $\Bar{C_t^{0,1,2}}$ are leveraged to compression image residual by contextual coding. Finally, we use a frame generator to reconstruct the decoded frame $\hat{x_t}$.

\subsection{Experimental Settings}

The proposed bi-directional video codec can serve as a plugin-in on any existing uni-directional video codec to compress the whole video sequence in I-B-P-B order. 

\subsubsection{Training Data.} We use Vimeo-90K\cite{xue2019video} training set. Videos are randomy cropped to $256\times256$ to train the bit-need model.

\subsubsection{Training Strategy.} We train the model to jointly optimize the rate-distortion cost : 
\begin{equation}
    L=\lambda\cdot MSE(x_t, \hat{x_t}) + R_F+R_x
\end{equation}
where $R_F$ and $R_x$ denote the bit rate in residual motion coding and contextual coding. $\lambda$ is set to 256. We use AdamW\cite{loshchilov2018fixing} and batch size of 4 to train the model.

\subsubsection{Benchmark.} We test the model in HEVC standard testing videos \cite{sullivan2012overview}. It contains 16 sequences including Class B, C, D, and E with different resolutions.

\subsubsection{Baseline.} In the paper we perform two experiment to compare with TCM \cite{sheng2021temporal} and H.265-HM \cite{HM}. TCM is the state-of-the-art learned P-frame codec. HM is the reference software for Rec. ITU-T H.265 | ISO/IEC 23008-2 High Efficiency Video Coding (HEVC). The intra period is set to 32.

\begin{figure*}[h]
    \centering
    \includegraphics[width=0.75\linewidth]{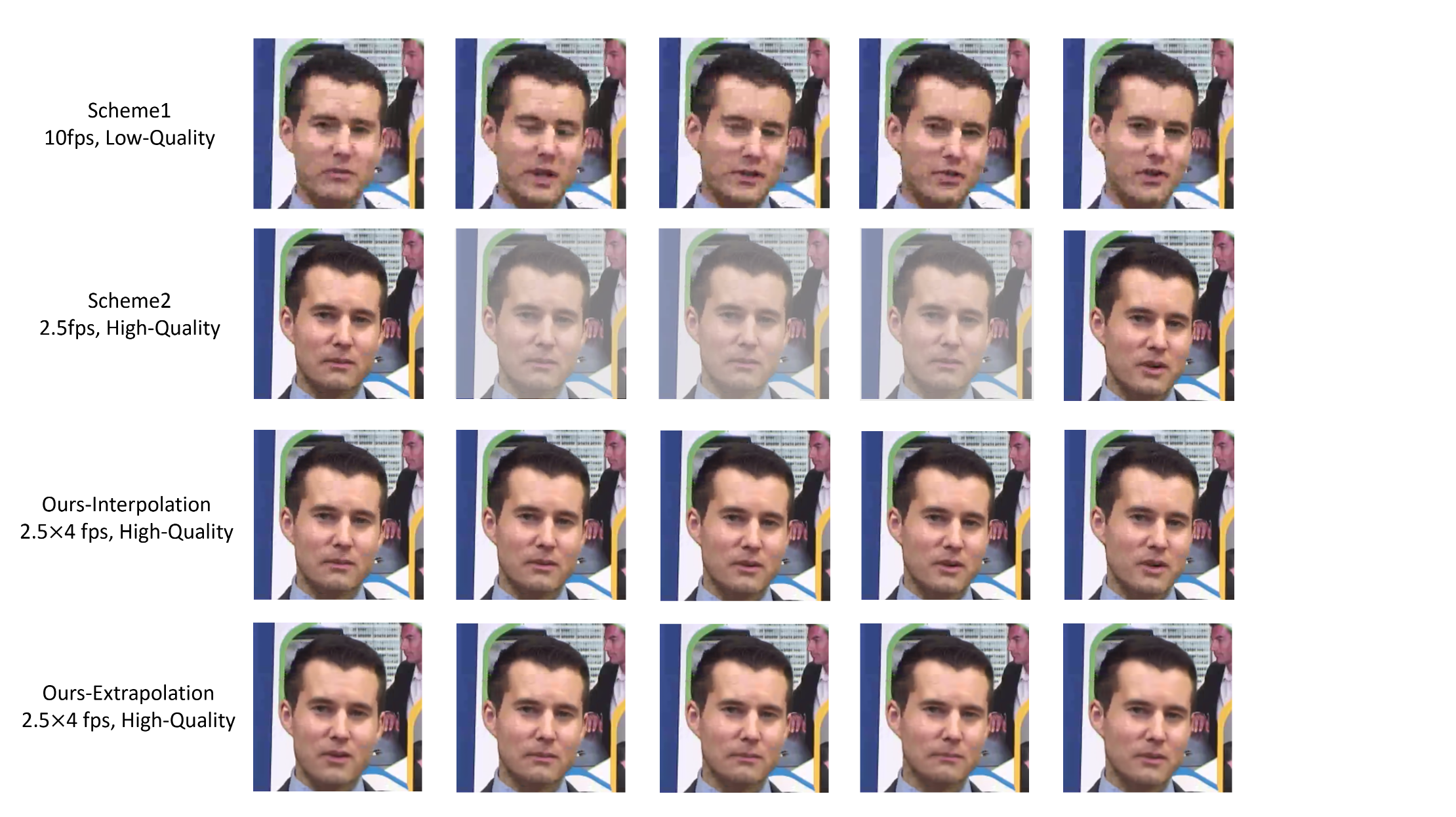}
    \caption{Jitter removal in bandwidth limited real-time communication. Best view in zoom in. Please see the video demo for better visual comparison.}
    \label{fig-sm9}
\end{figure*}

\begin{figure}[h]
    \centering
    \includegraphics[width=\linewidth]{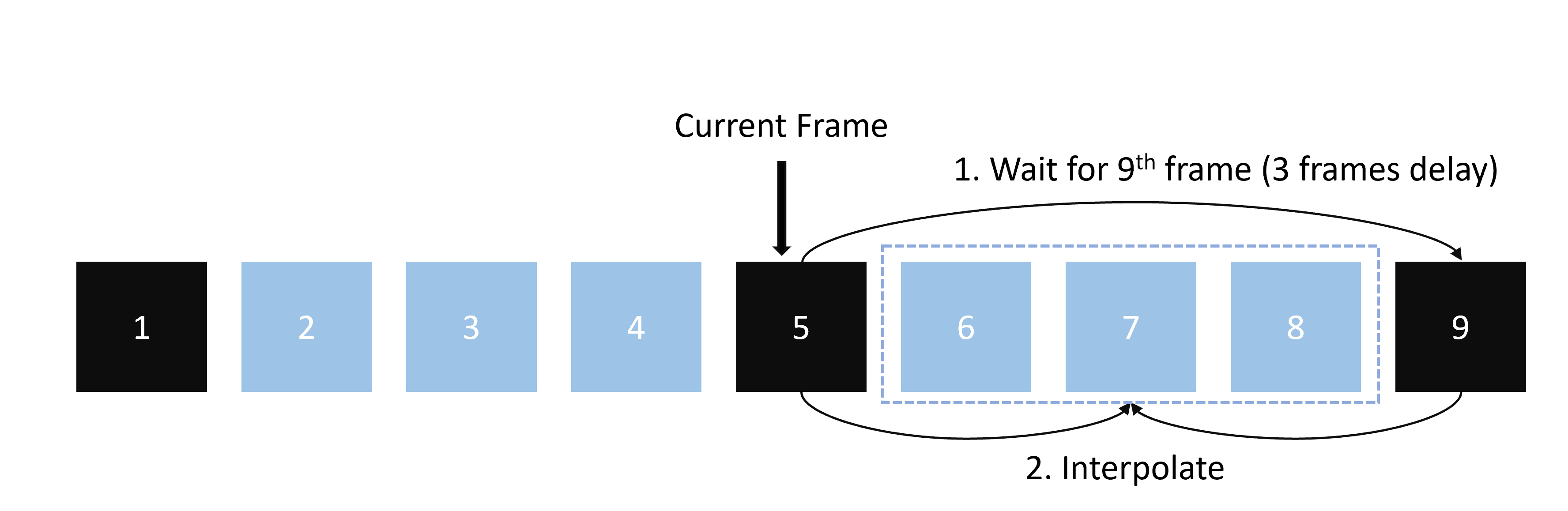}
    \caption{Jitter removal by interpolation.}
    \label{fig-sm10-1}
\end{figure}

\begin{figure}[h]
    \centering
    \includegraphics[width=\linewidth]{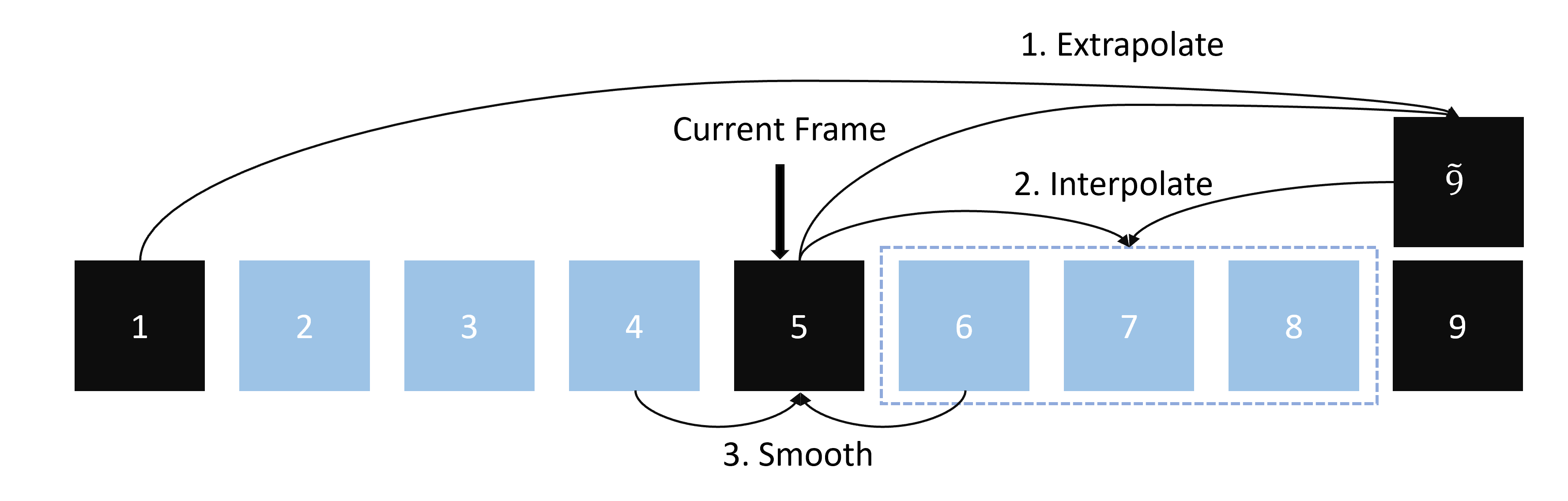}
    \caption{Jitter removal by extrapolation.}
    \label{fig-sm10-2}
\end{figure}
\subsection{Experiment Results}

We use TCM as the P-frame codec to compare with TCM in Fig.\ref{fig-sm4}. Our bit-free model is better or comparable with TCM on MS-SSIM, and outperforms TCM in HEVC D and E class on PSNR. In addition, the runtime on B-frame is about 9 times faster, from 500ms encoding and 252ms decoding time of TCM to 0ms encoding and 82ms interpolation time of the proposed model. Our bit-need model outperforms TCM in all datasets. 

We further use HM as the P-frame codec to compare the bit-need mode with HM in Fig.\ref{fig-sm5}. We test HM with the same I-B-P-B sequence order as ours, and results show that our bit-need model outperforms HM in all datasets. 

An example is shown in Fig.\ref{fig-sm8}, where we show the 17th and 18th frames of a compressed video. The proposed codec cost more bits in P-frame while it saves much more in B-frame. As a result, the overall performance in the whole sequence is better than the baseline (TCM).

\section{Application : Motion Effect}
We can use the proposed frame synthesis network to generate 
motion effects like motion blur and slow motion.

\subsection{Motion Blur Generation}
Motion blur is the apparent streaking of moving objects in a sequence of frames. In film or animation, simulating motion blur can lead to more realistic visuals. Simulating motion blur can also generate dataset for deblurring networks. 

Our interpolation model can be applied to generate such motion blur. We interpolate 31 frames between reference frames, and simply average them to generate the blurred image. As shown in Fig.\ref{fig-sm6}, the generated images are blurred in the fast moving regions while keep high quality in static regions.

\subsection{Slow Motion Generation}
Slow motion (slo-mo) is a common effect in film-making where time seems to be slowed down. Jiang et al. \cite{jiang2018super} proposed to use interpolation to generate slo-mo in videos. Our model can also be applied to generate slo-mo by interpolating multiple intermediate frames between given frames. Examples can be found in the video demos.

\section{Application : Jitter Removal}

In real-time communication (RTC), a long-standing problem is video jitter. There are many reasons for video jitter, such as bandwidth limitation and network transmission limitation. Here we mainly focus on the bandwidth limitation problem. Bandwidth limitation is a common problem in terminal devices. If the bandwidth is limited, the video must be compressed at a low rate, resulting in poor visual quality and user experience. Our model can be applied to remove such jitters to improve user experience.

\subsection{Baselines}

When the bandwidth is limited, in practice there are two straightforward solutions : 1) transmit high frame-rate video at low compression quality, and 2) transmit low frame-rate video at high compression quality. We denote them as "scheme1" and "scheme2" correspondingly.

\subsection{Methods}

We can transmit video at low frame-rate and high compression quality (scheme2), and use the proposed frame synthesis network to synthesis untransmitted frames. We propose two schemes to meet the need of different scenarios.

The first scheme (denoted as "Ours-Interpolation") utilize interpolation to synthesis untransmitted frames. As shown in Fig.\ref{fig-sm10-1}, after decoding the 5th frame, we wait for the 9th frame, and interpolate the 6th, 7th, 8th frames for display. This scheme can synthesis high quality frames, but it causes 3 frames delay in display.

The second scheme (denoted as "Ours-Extrapolation") utilize extrapolation to synthesis untransmitted frames without delay. As shown in Fig.\ref{fig-sm10-2}, after decoding the 5th frame, we use the 1st and 5th frames to extrapolate the 9th frame, and then interpolate the 6th, 7th, 8th frames. Since the non-linear motion and large temporal distance in low frame rate, the extrapolated frame may differ a lot from the raw frame. It leads to temporal inconsistency between the 4th and 5th (or between 8th and 9th) frames. To eliminate such limitation, we further use the 4th and 6th frame to interpolate the 5th frame for display. It can improve temporal smoothness in video.

\subsection{Results}

We show an example in Fig.\ref{fig-sm9}, where 5 frames in a video is shown. Scheme 1 suffers from the low visual quality and compression artifacts, and scheme 2 produces stuck playback. Compared with them, our methods is smoother and have higher quality, resulting in a better user experience. Please see the video demo for better visual comparison. Note that scheme 1 is not shown in video demo since the compression of video demo will influence the results.

\end{document}